\documentclass[a4paper,fleqn]{cas-dc-arxiv}

\usepackage[numbers]{natbib}

\usepackage{graphicx}
\usepackage{tabularx}
\usepackage{booktabs}
\usepackage{amssymb} 
\usepackage{textcomp}
\usepackage{pifont}
\usepackage{caption}
\usepackage{float}
\usepackage{makecell}
\usepackage{multirow}
\usepackage{placeins}
\usepackage{soul}
\usepackage{color, xcolor}
\sethlcolor{yellow}
\setul{1pt}{0.6pt}

\usepackage[linesnumbered,ruled,vlined]{algorithm2e}

\def\tsc#1{\csdef{#1}{\textsc{\lowercase{#1}}\xspace}}
\tsc{WGM}
\tsc{QE}
\tsc{EP}
\tsc{PMS}
\tsc{BEC}
\tsc{DE}
\begin{document}

\let\WriteBookmarks\relax
\def\floatpagepagefraction{1}
\def\textpagefraction{.001}
\shorttitle{BinoGen: Scaling egocentric binocular data for embodied visual perception and learning}
\shortauthors{Li et~al.}

\title [mode = title]{BinoGen: Scaling egocentric binocular data for embodied visual perception and learning}

\author[1,2,3]{Chunpeng Li}[
                        ]

\credit{Conceptualization, Data curation, Data analysis, Investigation, Visualization, Writing - original draft, Writing - review \& editing}

\author[2,3]{Ya-tang Li}[
                        orcid=0000-0003-2763-1534
                        ]
\cormark[1]
\ead{yatangli8@gmail.com}

\credit{Conceptualization, Supervision, Methodology, Investigation, Writing - original draft, Writing - review \& editing, Funding acquisition}                      

\cortext[cor1]{Corresponding author}
\affiliation[1]{organization={College of Biological Sciences, China Agricultural University},
            city={Beijing},
            country={China}}
\affiliation[2]{organization={Beijing Institute for Brain Research, Chinese Academy of Medical Sciences \& Peking Union Medical College},
            city={Beijing},
            country={China}}
\affiliation[3]{organization={Chinese Institute for Brain Research, Beijing (CIBR)},
            city={Beijing},
            country={China}}


\maketitle

\begin{abstract}
Embodied visual perception relies on temporally coherent visual experience accumulated through continuous engagement with the environment. However, collecting large-scale egocentric binocular observations together with dense annotations remains costly and difficult. Moreover, visual experience is shaped not only by the environment but also by the embodiment of the observer, including viewing height, field of view, binocular geometry, and motion through the scene. To address these challenges, we present \textit{BinoGen}, an automated framework for generating large-scale, embodiment-aware egocentric binocular visual experiences in indoor environments. BinoGen jointly models environmental and observer variation through generative scene synthesis, probabilistic object instantiation, appearance randomization, stochastic trajectory generation, and configurable binocular camera setups. The framework produces synchronized binocular videos together with dense multimodal supervision, including depth maps, optical flow, surface normals, semantic maps, object coordinates, and camera poses. Using BinoGen, we construct a dataset comprising more than 20 million annotated images for supervised learning. We demonstrate two complementary utilities of BinoGen. First, incorporating BinoGen data consistently improves real-world visual perception, including depth estimation, object detection, and video object tracking. Second, paired human-inspired and mouse-inspired observations from the same environments enable controlled investigation of how observer embodiment affects perceptual learning. Embodiment-specific adaptation substantially improves performance, while joint training enables a single model to perform competitively across both embodiments. Together, these results demonstrate that large-scale, controllable visual experience can improve embodied perception and provide a means to investigate how visual representations adapt and transfer across observer embodiments.
\end{abstract}

\section{Introduction}
\label{sec:intro}

Visual perception is critical to the survival and fitness of humans and other animals, providing essential information for behaviors such as scene understanding, spatial navigation, and decision-making. For embodied agents, perception is similarly grounded in temporally coherent visual experiences acquired through continuous interaction with the environment. These egocentric observations are inherently active and dynamic, shaped jointly by the visual environment and by the observer's configuration and movements. Training embodied perception tasks requires large-scale egocentric binocular observations in real-world environments with dense annotations, yet collecting such data remains expensive and labor-intensive. Importantly, 
static image datasets, such as ImageNet~\citep{deng_imagenet_2009}, VOC~\citep{everingham_pascal_2010}, and COCO~\citep{lin_microsoft_2014}, are typically captured from favorable viewing distances and viewpoints and lack the dynamic, egocentric nature required for embodied perception, making them unsuitable for training such tasks.

Large-scale real-world datasets have substantially advanced visual perception research by capturing both environments and observer-centered experiences.
Environment-centric datasets such as S3DIS~\citep{armeni_3d_2016}, ScanNet~\citep{dai_scannet_2017}, and Matterport3D~\citep{chang_matterport3d_2017} provide realistic 3D representations of indoor environments together with geometric and semantic annotations.
Complementing these resources, egocentric datasets capture observer-centered visual experience of humans~\citep{grauman_ego4d_2022} and other animals~\citep{6977451}, supporting embodied perception research in real-world settings.
However, these datasets lack dense multimodal annotations and synchronized binocular observations. 
Furthermore, they typically assume a fixed observer configuration, making it difficult to systematically investigate how embodiment shapes visual experience across environments.

Synthetic data offers a scalable alternative by enabling controllable generation of both environments and embodied visual experiences.
Recent advances in scene synthesis and simulation have dramatically improved environment diversity.
Methods such as ATISS~\citep{paschalidou_atiss_2021} and LEGO-Net~\citep{Wei_2023_CVPR} automatically generate realistic indoor layouts, whereas simulation platforms support photorealistic rendering with dense, consistent annotations across multiple modalities.
On the observer side, recent efforts have begun to generate embodied visual experiences.
EgoGen~\citep{li2024egogen} synthesizes egocentric observations through human--environment interaction, while UnrealZoo~\citep{Zhong_2025_ICCV} introduces diverse virtual creatures experiencing human-designed scenes. 
Although these advances substantially increase both environment and observer diversity, they primarily model generic or fictional agents and rarely incorporate biologically grounded embodiments, limiting our ability to study how embodied sensory configurations influence visual perception across species.

To address these limitations, we present \textit{BinoGen}, 
a scalable framework for generating embodiment-aware egocentric binocular visual experiences.
Unlike existing approaches that primarily vary the environment, BinoGen jointly models both environmental and observer diversity, allowing the same scene to be perceived through agents with different visual configurations.
The framework integrates automatic indoor scene synthesis, probabilistic object instantiation, appearance randomization, stochastic trajectory generation, configurable binocular embodiments, and physically based rendering into a unified pipeline.
From a shared scene representation, BinoGen generates paired observations across different embodiments while simultaneously producing dense multimodal annotations, including segmentation maps, depth maps, surface normals, object coordinates, optical flow, and camera poses. This pipeline yields a richly annotated synthetic video dataset of over 20 million binocular frames.

We investigate two complementary questions. First, can large-scale synthetic visual experience generated by BinoGen improve real-world embodied perception? Second, can paired observations generated under different embodiments in the same environment facilitate generalization across embodied agents? 
To answer these questions, we evaluate BinoGen across geometric, semantic, and temporal perception tasks.
Experimental results show that BinoGen consistently improves depth estimation, object detection, and video object tracking when combined with limited real-world supervision. Joint learning enables effective adaptation between human-like and mouse-like visual perception, allowing a single model to remain competitive across both embodiments.
The main contributions of this work are summarized as follows:
{
\begin{itemize}

    \item Introduce BinoGen, a scalable framework for automatically generating large-scale egocentric binocular visual experiences with both scene and observer diversity.
    \item Construct a large-scale, richly annotated synthetic video dataset of over 20 million binocular frames with dense multimodal annotations and paired observations from human-like and mouse-like embodiments.
    \item Demonstrate that BinoGen consistently improves real-world performance across geometric, semantic, and temporal embodied perception tasks, including depth estimation, object detection, and video object tracking. 
    \item Establish a controlled cross-embodiment evaluation using paired human-like and mouse-like observations, revealing substantial embodiment-induced domain shifts and demonstrating effective adaptation across embodiments.
    
\end{itemize}
}

\section{Related work}
\label{sec:related}

Research in embodied visual perception relies on a spectrum of data sources, which can be broadly categorized into real-world and synthetic data. 
Within each category, existing resources offer distinct trade-offs in realism, scalability, diversity, and annotation richness at both the environment and observer levels.

\subsection{Real-world environments and egocentric experience}
Large-scale real-world scene datasets capture the geometric, semantic, and appearance complexity of natural environments. 
Indoor scene datasets like ScanNet~\citep{dai_scannet_2017}, Matterport3D~\citep{chang_matterport3d_2017}, and S3DIS~\citep{armeni_3d_2016} provide realistic environments together with geometric and semantic annotations, and have become standard resources for scene understanding, reconstruction, and embodied navigation.
However, these datasets primarily characterize the environment itself rather than the visual experience of a particular embodied observer. 

Complementary to environment-centric datasets, egocentric datasets directly capture visual observations from body-mounted cameras.
EPIC-Kitchens~\citep{damen_scaling_2018}, Charades-Ego~\citep{sigurdsson_actor_2018}, and Ego4D~\citep{grauman_ego4d_2022} record human activities from wearable cameras and have boosted progress in action recognition and egocentric perception.
Several datasets have also extended egocentric capture beyond human observers.
EgoPet~\citep{bar2024egopet} and DogCentric Activity~\citep{6977451}, for example, collect videos from cameras mounted on animals, offering distinctive perspectives for studying embodied behaviors. 

However, these datasets have significant limitations for studying embodied visual perception. Designed primarily for behavioral analysis, they offer limited supervision for perceptual tasks and generally lack synchronized binocular video with dense annotations. Moreover, a body-mounted camera does not necessarily reproduce the animal's actual visual input, as camera placement, optical parameters, and field of view often differ from those of the biological observer. More fundamentally, environment properties and observer configurations are coupled at the time of data collection and cannot be independently varied afterward, making it difficult to systematically investigate how environment diversity and embodiment-specific observation jointly shape visual experience.

\subsection{Synthetic environments and egocentric experience}

Synthetic environments provide a scalable alternative to real-world data collection, while offering precise control over scene composition and automatically generated ground-truth annotations. Because observations are rendered from an underlying 3D scene representation, dense annotations can be generated automatically and remain precisely aligned with the corresponding images. On the one hand, resources such as ShapeNet~\citep{chang_shapenet_2015}, ModelNet~\citep{wu_3d_2015}, SceneNet RGB-D~\citep{mccormac_scenenet_2017}, and 3D-FRONT~\citep{fu20213d} support large-scale synthetic scene construction. On the other hand, generative methods including ATISS~\citep{paschalidou_atiss_2021}, DiffuScene~\citep{tang2024diffuscene}, and LEGO-Net~\citep{Wei_2023_CVPR} have improved the diversity and scalability of synthetic environments by automatically generating realistic indoor layouts.

Beyond static scene generation, simulation platforms such as Habitat~\citep{habitat19iccv} and Gibson~\citep{Xia_2018_CVPR} enable embodied agents to navigate interactive virtual environments and acquire egocentric visual observations, bridging synthetic environments and continuous visual experience. Building on this direction, recent methods have begun to synthesize observer-conditioned visual experiences. SceneDiffuser~\citep{Huang_2023_CVPR} models scene-aware human motion, while EgoGen~\citep{li2024egogen} generates egocentric data through simulated human--environment interactions, improving the realism of agent trajectories and egocentric observations. More recently, UnrealZoo~\citep{Zhong_2025_ICCV} has extended this direction by introducing diverse virtual embodiments within generated environments, broadening the range of observer configurations available for embodied learning.

Despite these advances, observer diversity and embodiment-dependent perception remain relatively underexplored. Existing approaches typically model observers as generic agents or virtual characters, with limited consideration of the perceptual characteristics of real biological species. Thus, it is difficult to disentangle how environmental variation and embodiment-specific visual input contribute to differences in visual experience and perceptual representations.

BinoGen addresses this gap by jointly modeling synthetic environments and species-specific embodiments within a unified visual-experience generation framework. The framework incorporates biologically informed observer configurations and trajectories, synchronized binocular rendering, and dense multimodal annotations, enabling the same environment to be rendered from different embodied perspectives. This design provides a scalable basis for generating and comparing embodiment-dependent visual experiences across species.

\section{Methods}
\label{sec:methods}
\begin{figure*}[ht!]
\begin{center}
	\centering
	\includegraphics[width=\textwidth]{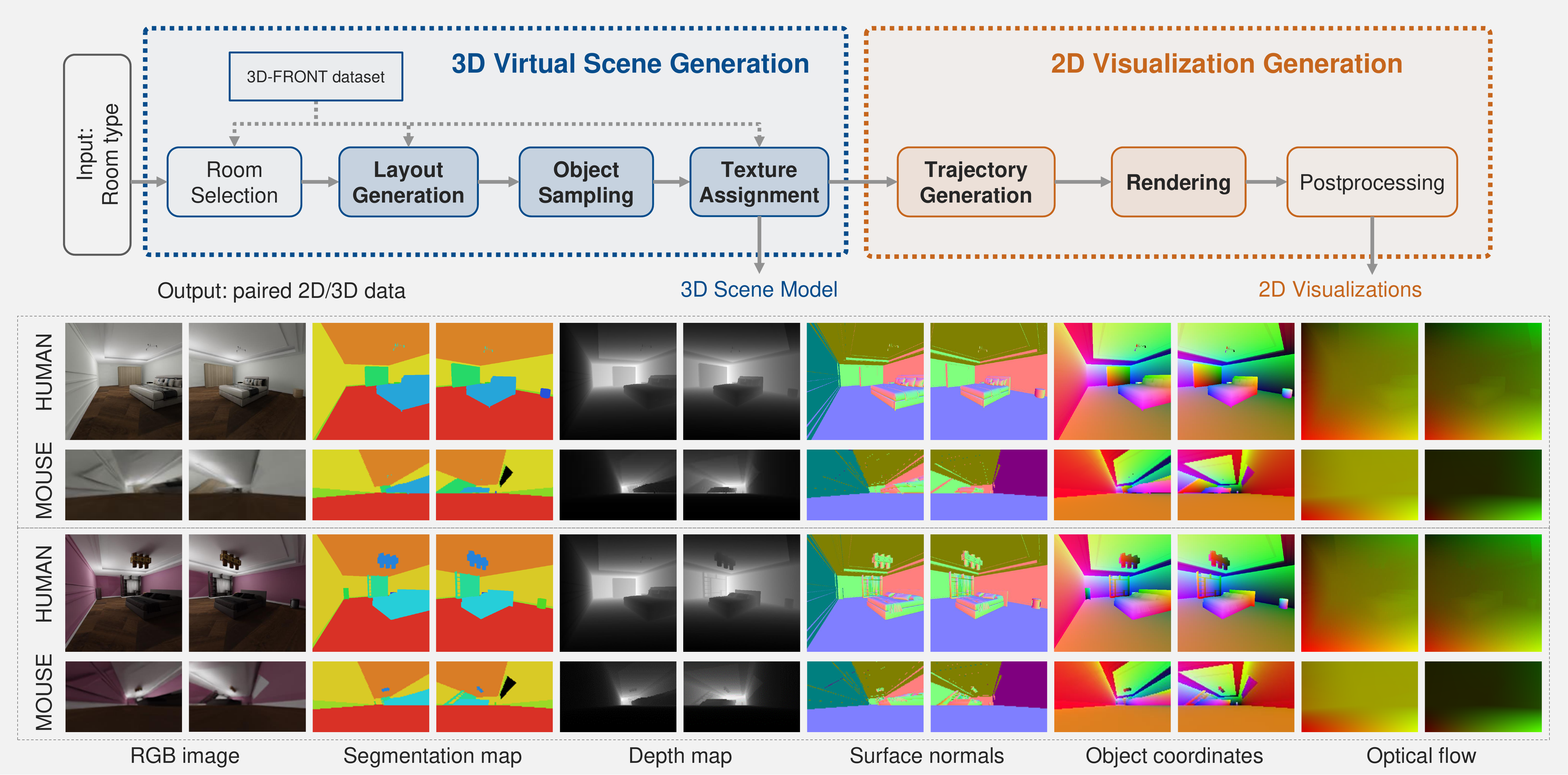}
	\caption{Overview of BinoGen. The framework for large-scale generation of egocentric binocular video. The top panel illustrates the generation pipeline, and the bottom panel shows example paired observations from different embodiments rendered from the same environment, accompanied by diverse annotations.}
	\label{fig.overview}
\end{center}%
\vspace{-5mm}
\end{figure*}

BinoGen is a fully automated framework for scalable synthesis of egocentric binocular videos in indoor environments (Fig.~\ref{fig.overview}).
The framework explicitly decouples the environment from the visual agent, enabling independent variation of scene content and observer properties.
We formalize a visual experience $\mathcal{V}$ as:
\begin{equation}
\mathcal{V} = \mathcal{R} \Big( \underbrace{(L,O,A)}_{\text{Environment}} \times \underbrace{(T,C)}_{\text{Visual Agent}} \Big),
\end{equation}
where $L$, $O$, and $A$ denote the layout, objects, and appearance of the environment, respectively; $T$ denotes the agent's trajectory; and $C$ denotes the embodiment configuration. 
As illustrated in Fig.~\ref{fig.overview} and Algorithm~\ref{alg:binogen}, BinoGen consists of three main stages.
First, it generates the geometry and appearance of an indoor environment.
Second, it generates a trajectory and binocular embodiment for each visual agent.
Finally, it renders synchronized binocular videos and aligned multimodal annotations.

\begin{algorithm}[htb]
	\caption{BinoGen: an automated data generation pipeline}
	\label{alg:binogen}
	\KwIn{room type $\tau$, 
			room dataset $\mathcal{D}_{r}$,
			object dataset $\mathcal{D}_{o}$,
			texture dataset $\mathcal{D}_{t}$,
			visual-agent configurations
			$\{\mathcal{C}^{(k)}\}_{k=1}^{K}$,
			sequence length $N$}
    \KwOut{environment $\mathcal{E}$ and 
			visual experiences $\mathcal{V}$}
	
	$\mathcal{V} \leftarrow \varnothing $\;
	Sample room  $r \sim \mathcal{D}_{r}|\tau$\;
	$L \leftarrow \texttt{LayoutGenerator}(r)$\;
	$O \leftarrow \varnothing$\;
	\ForEach{box $\ell_j \in L$}{
        $\mathcal{O}_{c_j} \leftarrow \texttt{CandidateObjects} (\mathcal{D}_{o},c_j)$\;
        $o_j^{*} \leftarrow \texttt{ProbabilisticSample} (\ell_j,\mathcal{O}_{c_j})$\;
		$O \leftarrow O \cup \{o_j^*\}$\;
	}
    $\mathcal{G} \leftarrow \texttt{AssembleGeometry} (r,L,O)$\;
    $A \leftarrow \texttt{GenerateAppearance} (r,\mathcal{D}_{t})$\;
    $\mathcal{E} \leftarrow (\mathcal{G},A)$\;
    $\mathbf{F}\leftarrow \texttt{CalculateSDF}(\mathcal{G})$\;
	
    $T_{\mathrm{planar}} \leftarrow \texttt{GenerateTrajectory}(\mathcal{E}, \mathbf{F}, N)$\;
    
    \For{$k\leftarrow1$ \KwTo $K$}{
        $T^{(k)} \leftarrow \texttt{ApplyEmbodiment}
        (T_{\mathrm{planar}}, \mathcal{C}^{(k)})$\;
        $\mathcal{V}^{(k)} \leftarrow
        \texttt{Render}(\mathcal{E}, T^{(k)}, \mathcal{C}^{(k)})$\;
    }
    
	\Return $\mathcal{E}, \mathcal{V}$
\end{algorithm}

\subsection{Environment generation}
The environment determines the visual content available to an agent.
The goal of environment generation is to automatically construct diverse 3D indoor scenes with plausible spatial configurations that approximate real-world indoor environments. 
Compared with manually designed scenes, BinoGen offers greater scalability and diversity, while maintaining realistic spatial layouts. 

Each virtual environment is represented by geometry and appearance: 
\begin{equation}
	\mathcal{E}
	=
    \underbrace{Geometry}_{\text{Layout+Objects}}+ \underbrace{Appearance}_{\text{Textures+Lighting}}.
\end{equation}
An environment is generated in three stages: layout generation, probabilistic object sampling, and appearance generation.
First, a learned scene-layout model generates semantically and spatially plausible object arrangements from curated indoor-scene datasets such as 3D-FRONT~\citep{fu20213d}. 
Next, concrete 3D object instances are probabilistically sampled from an asset library to populate the generated layout. 
Finally, textures and illumination are sampled to determine the appearance of the environment.

\subsubsection{Layout generation}
The first stage determines the scene layout by specifying the category, size, and spatial location of every object within an empty room. 

BinoGen provides a unified interface for different layout generators, supporting ATISS~\citep{paschalidou_atiss_2021}, LEGO-Net~\citep{Wei_2023_CVPR}, and DiffuScene~\citep{tang2024diffuscene}.
Because these methods employ compatible layout representations, they can be exchanged without modifying the remaining stages of the pipeline. 
For the dataset generated in this work, object layouts are generated using ATISS pretrained on 3D-FRONT~\citep{fu20213d}.
Given a room type, the layout model predicts a semantically valid object arrangement of objects represented by 3D bounding boxes.

\subsubsection{Probabilistic object sampling}
A generated layout specifies object categories and target bounding boxes but not the specific 3D object instances. 
Selecting the single best-fitting instance per bounding box can produce repeated scenes, whereas unconstrained random sampling may select objects that are poorly matched to the generated layout.
We therefore use a probabilistic, shape-aware sampling strategy to balance geometric compatibility and intra-category diversity.

For a target bounding box with object category $c$ and side lengths $\hat{\mathbf{s}}=(\hat w, \hat h, \hat d)$, the geometric discrepancy between the target box and each candidate object $i$ within the same category is computed as:
\begin{equation}
    D_i = (\hat w - w_i)^2 + (\hat h - h_i)^2  + (\hat d - d_i)^2.
\end{equation}
The candidate objects are ranked according to $D_i$. Let $N$ denote the number of candidates and $\{D_1\le D_2\le \cdots \le D_N\}$ denote the sorted distances. 
We assign each candidate a normalized rank:
\begin{equation}
    Q_i=\frac{\mathrm{rank}(D_i)}{N},
\end{equation}
where smaller values of $Q_i$ correspond to objects whose dimensions more closely match the target bounding box. 

Instead of always selecting the closest candidate, BinoGen introduces controlled stochasticity by sampling a percentile:

\begin{equation}
     u \sim \mathcal{N}_{[0,1]}(0, \sigma^2),
\end{equation}

The selected object $i^*$ is then determined by
\begin{equation}
    i^* = \arg\min_i{|Q_i-u|}.
\end{equation}
The parameter $\sigma$ controls the trade-off between geometric compatibility and object diversity.
A small value concentrates sampling near the best-matching candidate, whereas a larger value increases the probability of selecting alternative instances.

This strategy favors geometrically compatible objects while occasionally introducing controlled intra-category variation, thereby increasing scene diversity without substantially compromising spatial plausibility. 
Finally, the selected object is aligned with the target bounding box through translation and rotation to produce a visually coherent scene.

\subsubsection{Appearance generation}
To improve visual realism and prevent overfitting to specific appearances, BinoGen procedurally randomizes textures and illumination while preserving semantic consistency. A high-level interior style (e.g., \textit{modern}, \textit{classic}, or \textit{minimalist}) is first sampled from the 3D-FRONT dataset. Given the selected style, wall, floor, and other structural surface textures are randomly sampled from the corresponding material library. Whenever available, UV-preserving variants are used to ensure realistic surface mapping and to avoid rendering artifacts such as texture stretching or repetitive tiling.

\begin{figure}[htb!]
	\begin{center}

		\includegraphics[width=1\linewidth]{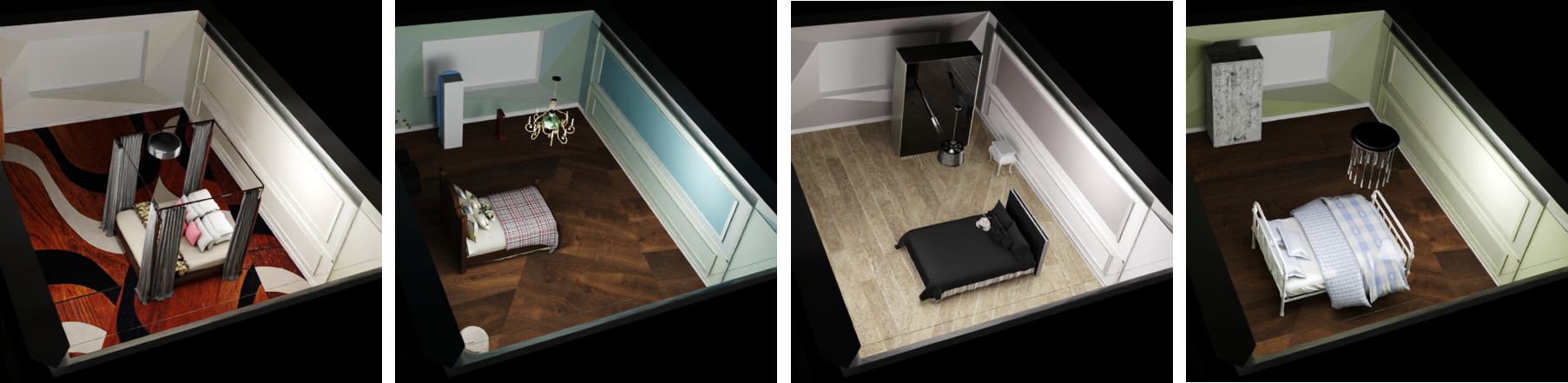}
	\end{center}
	\caption{Examples of generated texture assignments under different interior styles.}
	\label{fig.texture_div}
\end{figure}

Lighting is generated independently from geometry. Through stochastic variation of position, direction, intensity, and color temperature, BinoGen ensures that identical geometric configurations can produce diverse photometric conditions. As illustrated in Figure~\ref{fig.texture_div}, these randomized variations substantially increase visual diversity while maintaining semantic and spatial consistency, bringing the generated data closer to the statistical distribution of real-world indoor environments.  %shows examples of appearance variation generated from different interior styles.

\subsection{Visual agent generation}

The visual agent defines how an environment is explored and observed.
BinoGen instantiates each visual agent via two independent components: an egocentric trajectory and a binocular embodiment.
The trajectory determines the spatial and temporal sequence of observations, and the embodiment determines the binocular viewing geometry. These two components can be varied independently.

\subsubsection{Trajectory generation}

BinoGen adopts a stochastic kinematic navigation model to generate egocentric, physically plausible 6-DoF camera trajectories via an acceleration-sampling process that produces smooth head-like motion. 
Rather than learning an optimal navigation policy or explicitly mimicking species-specific locomotion, the model generates smooth and kinematically plausible trajectories with planar translation and three-axis camera rotation. This provides diverse, collision-free, and temporally coherent viewpoints for visual learning.

The pose at time $t$ consists of a position $\mathbf{p}_t\in\mathbb{R}^3$ and an orientation $\boldsymbol{\theta}_t=[\psi_t,\phi_t,\rho_t]^\top$, where $\psi_t$, $\phi_t$ and $\rho_t$ denote yaw, pitch, and roll, respectively.
Although only the resulting 6-DoF poses are used to place the camera, trajectory generation additionally maintains the linear velocity $\mathbf{v}_t$ and angular velocity $\boldsymbol{\omega}_t=[\dot{\psi}_t,\dot{\phi}_t,\dot{\rho}_t]^\top$ as internal motion states.
These velocities are updated by integrating the sampled accelerations, subsequently obtaining the next camera pose.

Each trajectory starts from a randomly selected corner $\mathbf{p}_0$, with another randomly selected corner $\mathbf{p}^*$ as the current navigation target. When the camera reaches the neighborhood of the target, defined by $\|\mathbf{p}_t-\mathbf{p}^{*}\|\le r_{\mathrm{switch}}$, a new target corner is sampled. This produces long trajectories that cover different regions of the environment.

\paragraph{Linear acceleration.} 
To mimic natural locomotion, the camera preferentially moves along its viewing direction with random directional perturbations, while the acceleration magnitude is independently sampled at each timestep:
\begin{equation}
\mathbf{a}_{\mathrm{lin},t}=a_t \cdot\frac{\mathbf v_t^{\mathrm{view}}+\lambda_\xi \boldsymbol\xi_t}{\left\|\mathbf v_t^{\mathrm{view}}+\lambda_\xi \boldsymbol\xi_t\right\|} ,
\label{eq:linear_acceleration}
\end{equation}
where $a_t\sim\mathcal U(0,A_{\mathrm{lin}}^{\max})$ denotes the acceleration magnitude, $\mathbf v_t^{\mathrm{view}}$ is the current viewing direction, $\boldsymbol\xi_t\sim\mathcal U([-0.5,0.5]^3)$ is a uniformly sampled directional perturbation, and $\lambda_\xi$ controls the perturbation strength.
The normalized sum in Eq.~\eqref{eq:linear_acceleration} determines only the direction of motion, whereas $a_t$ independently determines the acceleration magnitude. Translation is restricted to the horizontal plane.

The sampled acceleration is integrated to update the linear velocity and camera position: 
\begin{equation}
    \mathbf{v}_{t+1} = clip(\mathbf{v}_t+\mathbf{a}_{\mathrm{lin},t}\Delta t), \quad \mathbf{p}_{t+1} = \mathbf{p}_t+\mathbf{v}_{t+1}\Delta t,
\end{equation}
where $\operatorname{clip}(\cdot)$ preserves the velocity direction while restricting its magnitude. Trajectories are simulated at $10~\mathrm{Hz}$ with a maximum translational speed of $0.5~\mathrm{m/s}$ and a nominal acceleration limit of $5~\mathrm{m/s^2}$.

\paragraph{Angular acceleration.} 
The camera orientation is steered toward the current navigation target while retaining stochastic head-like variations. The target vector $\mathbf{p}^{*}-\mathbf{p}_t$ is converted to egocentric spherical coordinates: $\psi_{\mathrm{goal}}$ and $\phi_{\mathrm{goal}}$.
The normalized yaw and pitch offsets are defined as:
\begin{equation}
\Delta\psi_t=\frac{C_\psi(\psi_{\mathrm{goal}}-\psi_t)}{\psi_{\mathrm{clip}}},
\qquad
\\%
\Delta\phi_t=\frac{\phi_{\mathrm{goal}}-\phi_t}{\phi_{\max}-\phi_{\min}},
\end{equation}
where $\mathcal{C}_{\psi}$ first resolves the angular periodicity of the yaw error and then clips it to $\pm90^\circ$. The clipped error is divided by $\psi_{\mathrm{clip}}=90^\circ$, yielding $\Delta\psi_t\in[-1,1]$.
These normalized offsets serve as the means of Gaussian acceleration models for angular accelerations:
\begin{align}
\alpha_{\psi,t}&=A_\psi\epsilon_{\psi,t}, \qquad \epsilon_{\psi,t}\sim\mathcal{N}(\Delta\psi_t,\sigma_\psi^2),
\\
\alpha_{\phi,t}&=A_\phi\epsilon_{\phi,t}, \qquad \epsilon_{\phi,t}\sim\mathcal{N}(\Delta\phi_t,\sigma_\phi^2).
\end{align}

Roll accelerations are sampled independently using a damped stochastic process:
\begin{equation}
\alpha_{\rho,t}=A_\rho\epsilon_{\rho,t}, \qquad \epsilon_{\rho,t}\sim\mathcal{N}\left(-\frac{\dot{\rho}_t}{\rho_{\max}-\rho_{\min}},\sigma_\rho^2\right),
\end{equation}
which stabilizes camera orientation while preserving natural head motion. 

The sampled angular accelerations are collected as
$\boldsymbol{\alpha}_t=
[\alpha_{\psi,t},\alpha_{\phi,t},\alpha_{\rho,t}]^\top$
and integrated to update the orientation:
\begin{equation}
        \boldsymbol\omega_{t+1} = \mathcal{C}_{\omega}(\boldsymbol\omega_t+\boldsymbol\alpha_t\Delta t), \quad \boldsymbol\theta_{t+1} = \mathcal{C}_{\theta}(\boldsymbol\theta_t+\boldsymbol\omega_{t+1}\Delta t),
\end{equation}
where $\mathcal{C}_{\omega}$ limits $\dot{\psi}_t \in [-18^\circ/\mathrm{s}, 18^\circ/\mathrm{s}]$ and $\mathcal{C}_{\theta}$ enforces $\phi_t \in [-30^\circ,10^\circ]$ and $\rho_t \in [-30^\circ,30^\circ]$.
$A_\psi$, $A_\phi$ and $A_\rho$ are set to $36$, $40$, and $40~^\circ/\mathrm{s^2}$, respectively.

\paragraph{Collision avoidance.}
To prevent collisions, we compute a 2D signed distance field (SDF) on the ground plane representing the distance from the camera position to the nearest object boundary. 
Let $F(\mathbf{p})$ denote the distance from position $\mathbf{p}$ to the nearest obstacle boundary.
A position is considered unsafe when $F(\mathbf{p}_t)<r_{\mathrm{safe}}$.
In this case, the SDF gradient
\begin{equation}
    \mathbf{g}_t
    =
    \frac{
        \nabla F(\mathbf{p}_t)
    }{
        \left\|
        \nabla F(\mathbf{p}_t)
        \right\|_2
    }
    \label{eq:sdf_gradient}
\end{equation}
defines a local direction away from the nearest obstacle.
Whenever a potential collision is detected, both linear and angular accelerations are resampled along the steepest SDF gradient, steering the trajectory away from nearby obstacles. 

\paragraph{Shared trajectory across embodiments.}
The trajectory generator is shared across visual agents.
This design separates motion variation from embodiment variation.
For controlled comparisons, different visual agents can follow a matched planar path while using embodiment-specific viewing heights and camera configurations.
Compared with learning–based trajectory generation (e.g., SceneDiffuser~\citep{Huang_2023_CVPR}), this stochastic kinematic model generates collision-free trajectories with substantially higher efficiency.
This design allows embodiment-related perceptual differences to be studied independently of behavioral differences.

\subsubsection{Binocular embodiment}

The binocular embodiment defines how a visual agent observes the environment.
Unlike datasets that assume a fixed camera configuration, BinoGen treats the binocular configuration as an explicit embodiment model that can be adjusted to represent different visual systems.

In this work, we instantiate human-like and mouse-like configurations.
Their configurations differ in visual resolution, field of view, view angle, binocular overlap, interocular distance, focal length, and viewing height ( Table\ref{tab:config}).
These parameters capture geometric properties of visual acquisition and determine the visual domain experienced by each agent.

For controlled cross-embodiment generation, visual agents share the same environment and planar trajectory. Only the embodiment configuration is changed.
This produces paired visual sequences in which environmental content and motion timing are controlled while the visual input varies according to the observer's embodiment.

\begin{table}[ht]
\caption{Embodiment configurations for human and mouse visual agents.}
\begin{tabular}{lcc}
\hline
                        & \textbf{Human}  & \textbf{Mouse} \\ \hline
\textbf{Image resolution}        & 512$\times$512  & 150$\times$90\\
\textbf{Horizontal FOV}          & 90$^\circ$      & 150$^\circ$   \\
\textbf{Vertical FOV}            & 90$^\circ$      & 90$^\circ$    \\
\textbf{Azimuth}                 & $\pm$15$^\circ$ & $\pm$60$^\circ$\\
\textbf{Elevation}               & 0$^\circ$       & 40$^\circ$    \\
\textbf{Binocular overlap}       & 60$^\circ$      & 15$^\circ$    \\
\textbf{Interocular distance}    & 70 mm           & 10 mm\\ 
\textbf{Focal length}            & 22 mm           & 2 mm  \\
\textbf{Camera height}           & 150 cm          & 5 cm  \\ \hline
\end{tabular}
\label{tab:config}
\end{table}

\subsection{Visual experience rendering}

Given an environment and a visual agent, BinoGen renders a temporally coherent binocular sequence.
Rendering incorporates ray tracing, texture mapping, and global illumination to generate photorealistic binocular videos. 
Each rendered frame is accompanied by synchronized annotations, including RGB images, depth maps, segmentation maps, optical flow, surface normals, and object coordinates.

To support large-scale dataset construction, rendering is parallelized across multiple GPUs, enabling efficient generation of millions of annotated frames. 
Each rendering worker maintains a persistent Blender process and reuses the loaded scene state across rendering tasks. This warm-start strategy avoids repeated initialization and redundant asset loading, substantially improving rendering efficiency.
The resulting dataset provides temporally coherent stereo videos with dense multimodal supervision, suitable for training and evaluating embodied perception models.

\section{Dataset analysis}
\label{sec:dataset}
To characterize the properties of the dataset generated by BinoGen, we evaluate the dataset along four dimensions: dataset scale, environmental diversity, visual-experience diversity, and generation efficiency.
We first summarize the overall dataset statistics, then quantify diversity in scene layouts, object instances, appearance, trajectories, and visual embodiments. Finally, we evaluate the computational efficiency of the rendering pipeline for large-scale dataset construction.

\subsection{Dataset statistics}
Using this pipeline, we constructed a large-scale binocular egocentric dataset comprising 20,000 unique indoor scenes spanning four room categories: bedroom, living room, dining room, and library (5,000 scenes each). For each scene, BinoGen generates two synchronized binocular videos, producing 40,000 videos in total. Each video contains 500 frames rendered at 10 FPS, yielding 20 million annotated images. Every frame is accompanied by perfectly aligned depth maps, optical flow, surface normals, segmentation masks, object coordinates, and camera poses. These annotations provide synchronized geometric, semantic, and temporal supervision for a broad range of embodied perception tasks.

\subsection{Diversity of environments}

We evaluated the diversity of environments from three complementary perspectives: layout diversity, object diversity, and appearance diversity. Layout diversity captures variation in the spatial organization of indoor scenes, object diversity measures variation among concrete object instances within semantic categories, and appearance diversity characterizes photometric variation induced by textures and illumination.

\subsubsection{Layout diversity}

To quantify structural variation among generated scenes, we randomly sampled 10,000 scene pairs from 5,000 independently generated layouts for each room category. For each pair, we computed the mean 3D intersection-over-union (IoU) between bounding boxes of objects belonging to the same semantic categories. The resulting IoU distribution provides a measure of structural similarity, with lower IoU values indicating greater structural diversity.

\begin{figure}[htb!]
	\begin{center}

		\includegraphics[width=0.9\linewidth]{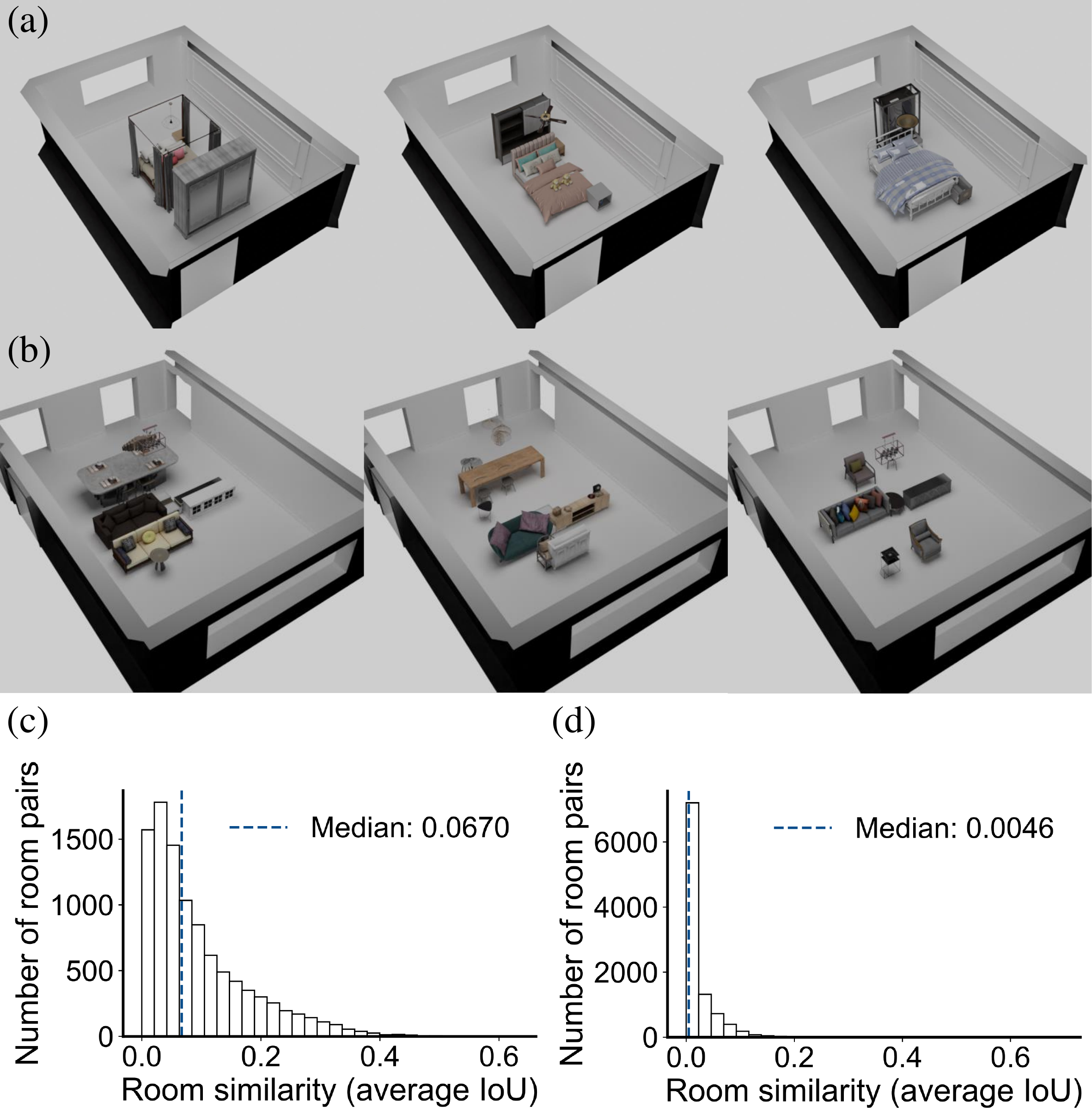}
	\end{center}
	\caption{Layout diversity. (a,b) Representative bedroom and living-room layouts generated by BinoGen. (c,d) Distributions of the mean 3D IoU across 10,000 independently generated layout pairs for bedrooms and living rooms.}
	\label{fig.layout_div}
\end{figure}

Representative bedroom and living-room layouts are shown in Fig.~\ref{fig.layout_div}. The corresponding IoU distributions demonstrate substantial variation in object placement and spatial organization, indicating that the layout generator produces diverse configurations rather than minor perturbations of a common template.

\subsubsection{Object diversity}

\begin{figure}[htb!]
	\begin{center}

		\includegraphics[width=\linewidth]{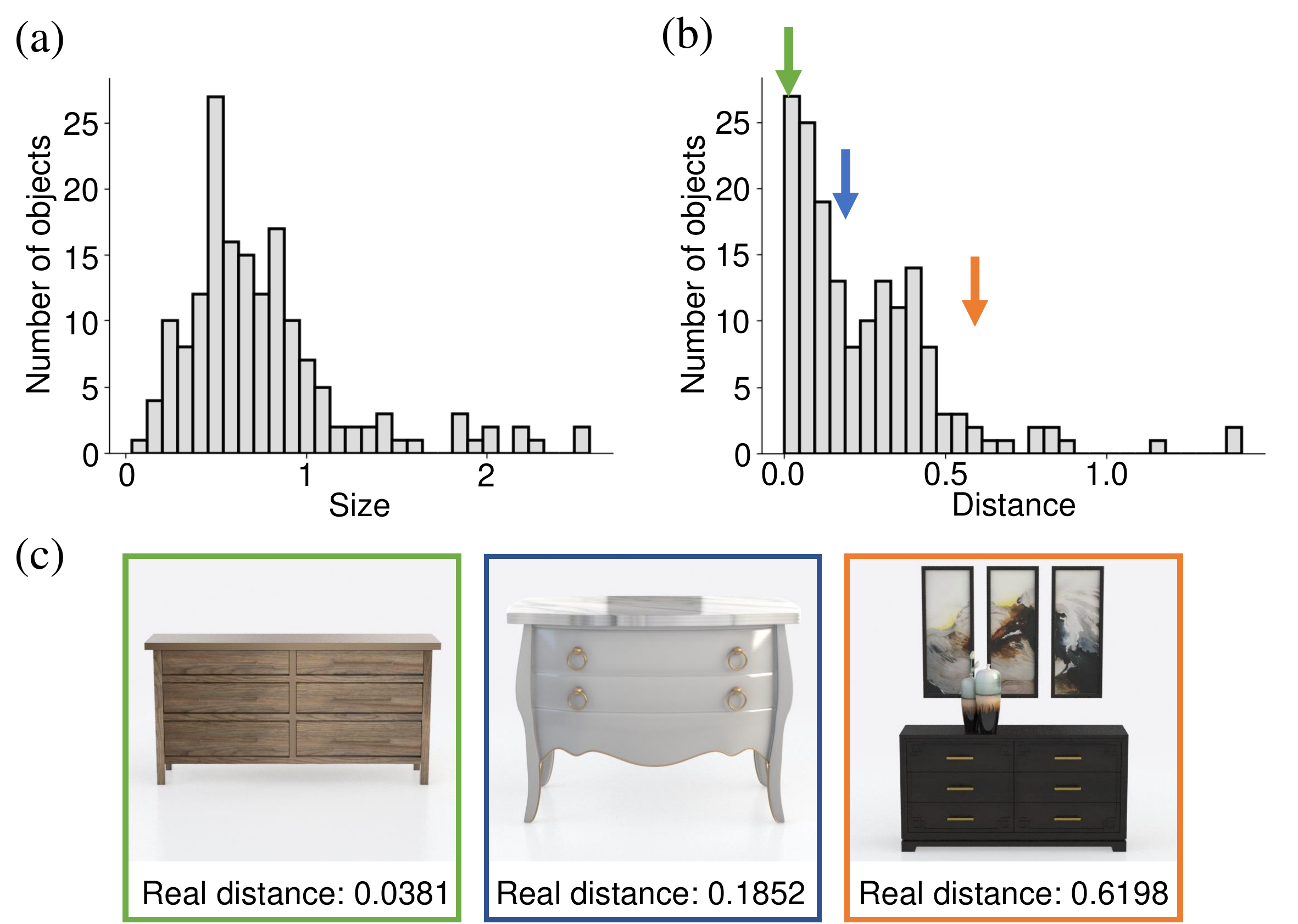}
	\end{center}
	\caption{Object diversity produced by probabilistic sampling. (a) Distribution of cabinet sizes. (b) Distribution of geometric distances to a target bounding box. (c) Representative sampled object instances.}
	\label{fig.size_div}
\end{figure}

Beyond layout variation, BinoGen increases intra-category diversity through probabilistic object sampling. For each generated bounding box, candidate object instances are selected from the asset library according to geometric similarity to the target and then sampled probabilistically rather than deterministically selecting the closest match (Fig.~\ref{fig.size_div}). This strategy preserves compatibility with the generated layout while allowing different object instances with similar dimensions to populate otherwise identical layouts.

The combination of layout variation and stochastic object sampling enables BinoGen to synthesize a large number of distinct indoor scenes while retaining the semantic and spatial structure specified by the generated layouts.

\subsubsection{Appearance diversity}

\begin{figure}[h]
	\begin{center}

		\includegraphics[width=\linewidth]{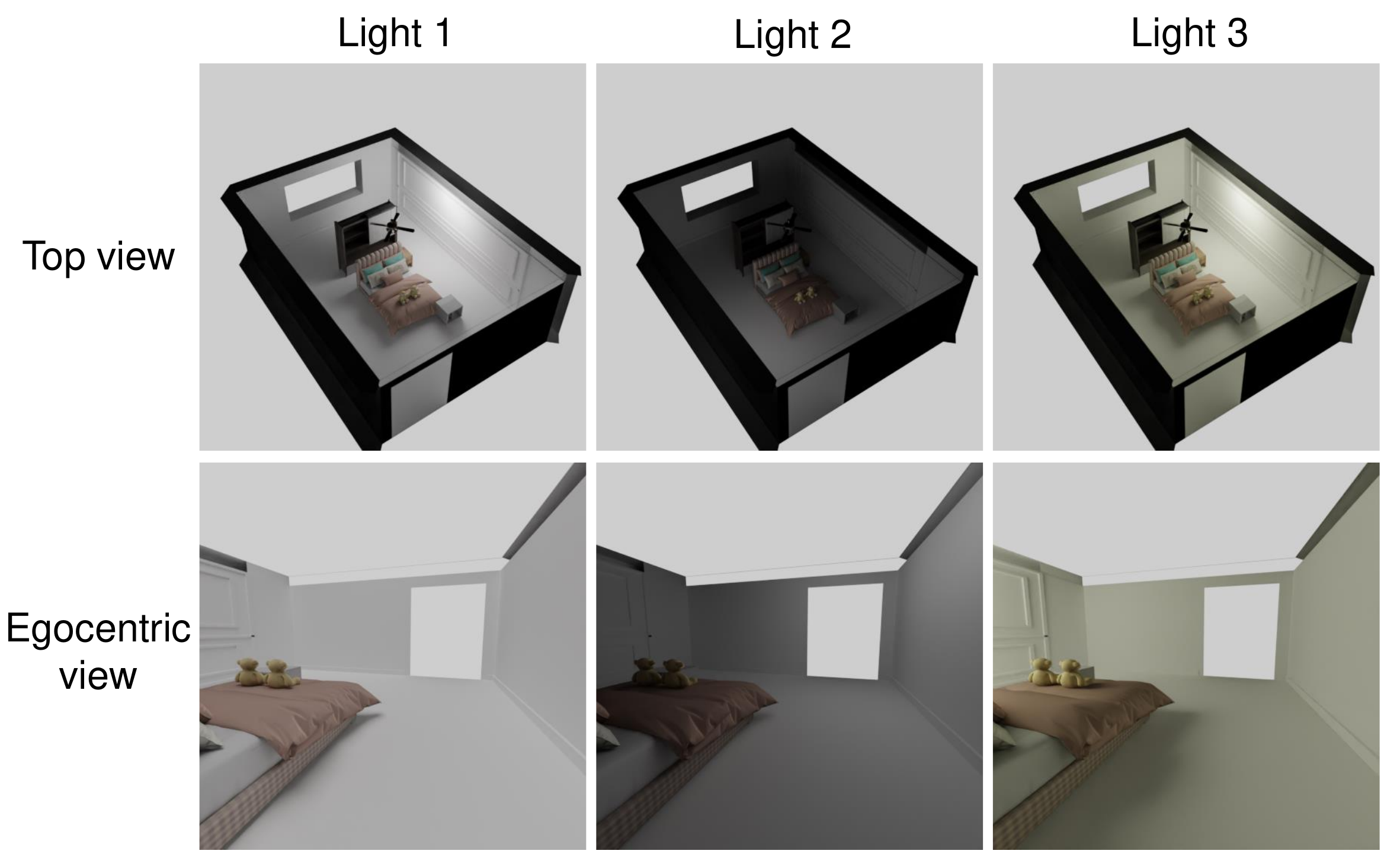}
	\end{center}
    \caption{Lighting diversity. The same scene rendered under different illumination conditions.}
	\label{fig.light_div}
\end{figure}

BinoGen further introduces appearance diversity by independently randomizing textures and illumination while preserving the underlying scene geometry. Lighting conditions were varied in terms of illumination intensity, lighting direction, and color temperature. Thus, the same scene can exhibit substantially different visual appearances under different lighting configurations (Fig.~\ref{fig.light_div}). Such photometric variation exposes visual models to a broader range of appearance conditions and can improve robustness to illumination changes encountered in real-world environments.

\subsection{Diversity of visual experience}

Environmental diversity alone does not determine the visual experience available to an agent. The same environment can produce substantially different observations depending on how it is explored and the visual system through which it is observed. BinoGen introduces two additional sources of experience-level variation: trajectory diversity and embodiment diversity.

\subsubsection{Trajectory diversity}

\begin{figure}[htb!]
	\begin{center}

		\includegraphics[width=\linewidth]{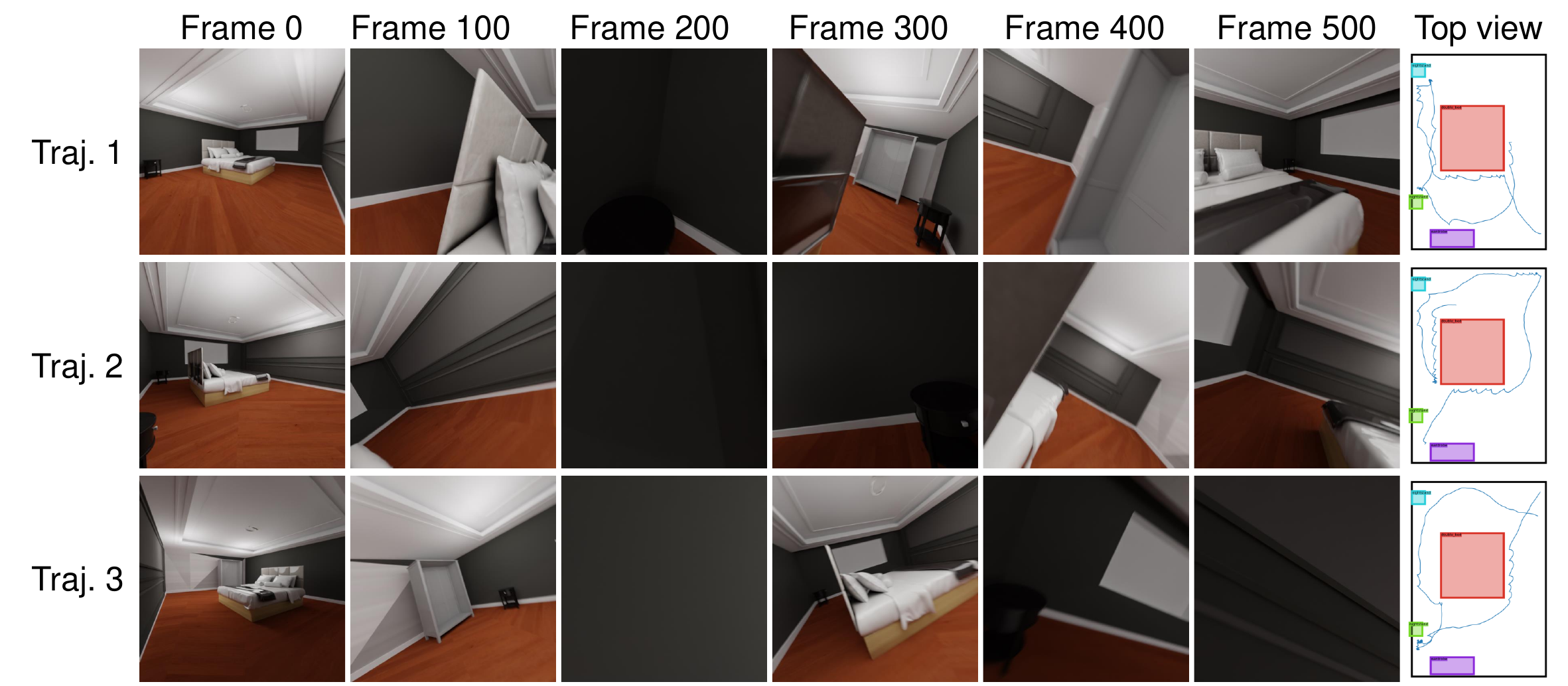}
	\end{center}
	\caption{Trajectory diversity. Different camera trajectories within the same environment produce distinct egocentric visual experiences.}
	\label{fig.traj_div}
\end{figure}

Different egocentric trajectories alter the sequence of viewpoints through an environment and provide distinct visual experiences. Fig.~\ref{fig.traj_div} shows representative trajectories within the same environment, which generate different sequences of visual observations despite identical scene geometry.

Because trajectories can be independently generated for a given environment, a finite set of scenes can yield a substantially larger number of unique visual experiences. This provides an additional axis of scalability beyond the diversity of the underlying scene library.

\subsubsection{Embodiment diversity}

BinoGen introduces an additional source of visual-experience diversity by varying the embodiment of the visual agent. The embodiment determines how the environment is projected onto the agent's visual system, including viewing geometry, field of view, and binocular configuration.

\begin{figure}[htb!]
	\begin{center}

		\includegraphics[width=\linewidth]{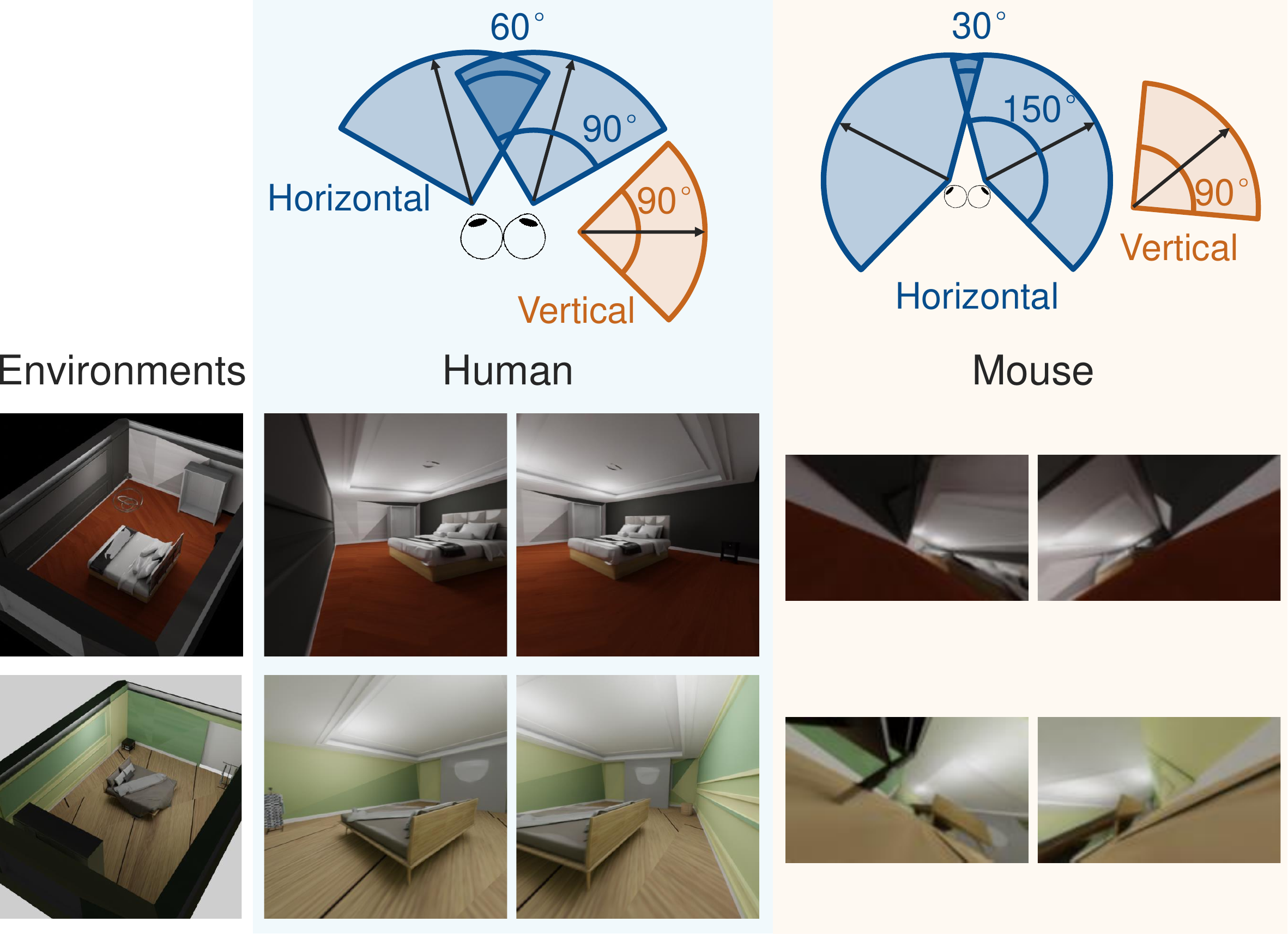}
	\end{center}
	\caption{Embodiment diversity. Observations from human-viewpoint and mouse-viewpoint configurations within the same environment produce distinct egocentric visual experiences.}
	\label{fig.embd_div}
\end{figure}

We instantiate two embodiment configurations inspired by human and mouse vision. Human-viewpoint videos were rendered at a resolution of $512\times512$ pixels using a 22~mm focal length and a horizontal field of view (FOV) of 90$^\circ$, with the optical axis aligned to the horizon. Mouse-viewpoint videos were rendered at $150\times90$ pixels using a 2~mm focal length and a 150$^\circ$ horizontal FOV, with the optical axis oriented $40^\circ$ above the horizon. The two configurations further differ in viewing height, binocular geometry, and other camera parameters, as summarized in Table~\ref{tab:config}. For controlled cross-embodiment generation, both visual agents observe the same environment while following the same planar trajectory. Figure~\ref{fig.embd_div} compares the resulting paired human-viewpoint and mouse-viewpoint observations.

These embodiment-specific configurations substantially alter the visual representation of the same environment, resulting in differences in object scale, visible spatial extent, and binocular relationships. Because scene content and motion are held constant, the resulting paired observations isolate changes in visual experience attributable to embodiment. This controlled cross-embodiment variation provides the basis for the cross-species adaptation experiments in Sec.~\ref{species}.

\subsection{Generation efficiency}

BinoGen is designed for efficient large-scale dataset generation through parallelized rendering and persistent scene initialization. Rendering was performed in Blender using the physically based renderer Cycles via Kubric~\citep{Greff_2022_CVPR}. To increase rendering throughput, we employed distributed multi-instance rendering with two Blender processes per GPU on four NVIDIA RTX 4090 GPUs. Each rendering process maintains a persistent Blender instance and reuses the loaded scene state across rendering tasks, avoiding repeated scene initialization and asset loading.

For each generated scene, BinoGen renders synchronized binocular videos together with dense multimodal annotations, including RGB images, depth maps, optical flow, surface normals, segmentation masks, object coordinates, and camera poses. This unified rendering procedure ensures that all annotations remain spatially and temporally aligned with the corresponding visual observations.

\begin{figure}[htb!]
	\begin{center}

		\includegraphics[width=\linewidth]{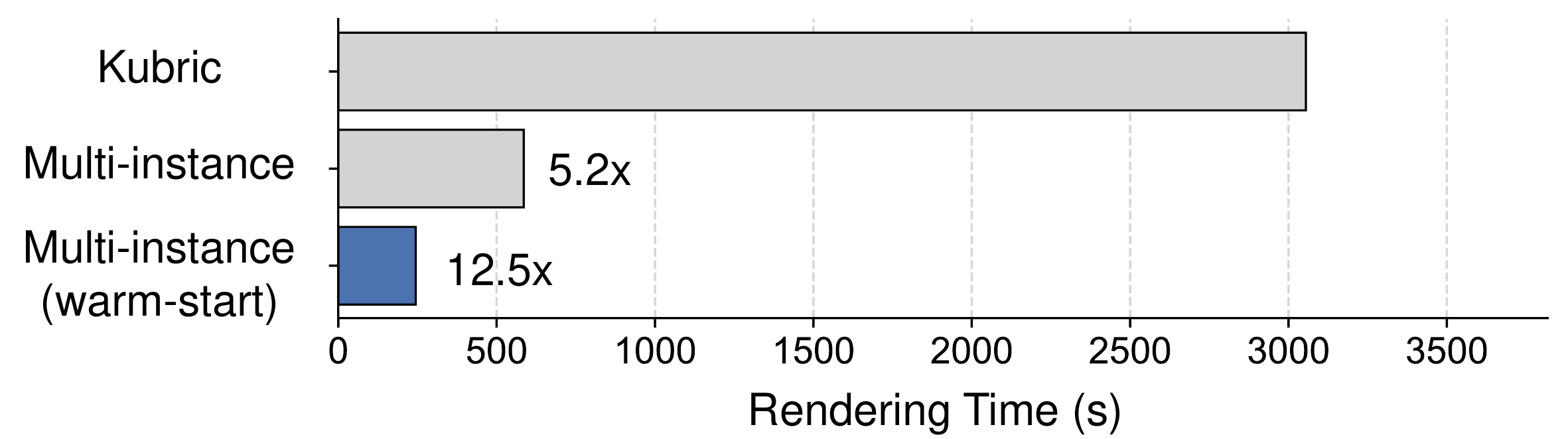}
	\end{center}

    \caption{Rendering time comparison between the original Kubric pipeline and our multi-instance rendering implementation.}
	\label{fig.time}
\end{figure}

To quantify the efficiency gains, we compared the original Kubric rendering pipeline with our parallel multi-instance Blender implementation using identical 500-frame sequences at a resolution of $512\times512$ pixels. As shown in Fig.~\ref{fig.time}, multi-instance rendering substantially reduces rendering time relative to the original pipeline. Combined with persistent scene initialization, this implementation enables BinoGen to efficiently scale video synthesis to tens of millions of annotated frames.

\subsection{Comparison with existing datasets}

BinoGen combines large-scale binocular egocentric videos, multiple visual embodiments, and dense multimodal ground truth within a fully automated generation framework. Table~\ref{tab.scene_cmp} compares BinoGen with representative indoor scene datasets, while Table~\ref{tab.gt_cmp} compares its visual and annotation capabilities with existing egocentric video datasets.

Compared with existing indoor scene datasets, BinoGen provides a substantially larger collection of procedurally generated scenes while retaining explicit 3D scene geometry. In addition, its procedural generation framework enables systematic variation of layouts, object instances, and appearances rather than relying solely on manually curated scenes.

Compared with existing egocentric video datasets, BinoGen uniquely combines binocular observations with human-like and mouse-like visual embodiments and dense per-frame geometric, semantic, and temporal annotations. These properties enable controlled investigation of embodied visual perception, including both within-embodiment learning and cross-embodiment adaptation.

\setlength{\tabcolsep}{3pt}
\begin{table}[htb!]
	\centering
    \caption{Comparison of indoor scene datasets.} 
	\small
	
	\begin{tabularx}{.45\textwidth}{lccc} 
		
		\midrule[0.5pt]
		\textbf{Dataset}     & \makecell[c]{\textbf{Num.~of}\\ \textbf{scenes}} & \makecell[c]{\textbf{Object}\\ \textbf{categories}} & \textbf{3D~Ground Truth}  \\ 
		\midrule[0.5pt]
		\text{ScanNet}  & 707                     & 17                          & Estimated~Mesh  \\
		\text{3D-FRONT} & 7K                      & 45                        & Mesh            \\
        \midrule[0.5pt]

        \textbf{BinoGen (Ours)} & \textbf{20K} & \textbf{45} & \textbf{Mesh} \\
		\midrule[0.5pt]
	\end{tabularx}
	\label{tab.scene_cmp}
\end{table}

\begin{table*}[htb!]
    \centering
    \caption{Comparison between BinoGen and representative egocentric datasets.}

    \small
    \newcolumntype{Y}{>{\centering\arraybackslash}X}
    \setlength{\tabcolsep}{2pt}
    \begin{tabularx}{1\linewidth}{lccYYYccYYY} 
        \midrule[0.5pt]
        \textbf{Dataset} & \makecell[c]{\textbf{Num. of} \\ \textbf{Videos}} & \makecell[c]{\textbf{Avg.} \\ \textbf{Length}} & \makecell[c]{\textbf{Human} \\ \textbf{view}} & \makecell[c]{\textbf{Animal} \\ \textbf{view}} & \makecell[c]{\textbf{Segmen} \\ \textbf{-tation}} & \textbf{Depth} & \makecell[c]{\textbf{Optical} \\ \textbf{flow}} & \makecell[c]{\textbf{Surface} \\ \textbf{normal}} & \makecell[c]{\textbf{Object} \\ {\textbf{coordinate}}} & \makecell[c]{\textbf{Scene} \\ \textbf{geometry}} \\ 
        \midrule[0.5pt] 
        UT Ego \cite{6247820}
        & 10 & 4 h & Mono & \ding{55} & \ding{55} & \ding{55} & \ding{55} & \ding{55} & \ding{55} & \ding{55} \\
        DogCentric Activity \cite{6977451}
        & 208 & 5 s & \ding{55} & Mono & \ding{55} & \ding{55} & \checkmark & \ding{55} & \ding{55} & \ding{55} \\
        Ego4D \cite{grauman_ego4d_2022}
        & 10K & 5 min & \mbox{Bino (few)} & \ding{55} & \ding{55} & \ding{55} & \ding{55} & \ding{55} & \ding{55} & Few \\
        \midrule[0.5pt]

        \textbf{BinoGen (Ours)}
& \textbf{40K} & \textbf{50 s} & \textbf{Bino} & \checkmark & \checkmark & \checkmark & \checkmark & \checkmark & \checkmark & \checkmark \\
        \midrule[0.5pt]
    \end{tabularx}
    \label{tab.gt_cmp}
\end{table*}

\section{Experiments} 
\label{sec:app_exp}
We evaluate BinoGen from three complementary perspectives.
First, we assess its practical utility as a scalable source of supervision for real-world embodied perception. We consider three tasks spanning geometric, semantic, and temporal aspects of visual understanding: depth estimation, object detection, and video object tracking.
Second, we investigate the effect of embodiment shift and the ability to adapt to a target embodiment. Using paired observations rendered along the same trajectories in the same environments under human-inspired and mouse-inspired binocular configurations, we examine how changes in visual embodiment affect the transferability of learned representations and whether target-embodiment experience can bridge the resulting domain gap.
Third, we investigate joint learning across embodiments. By training models jointly on human-viewpoint and mouse-viewpoint visual experiences, we ask whether a single model can learn representations that remain effective across substantially different visual domains. Together, these experiments characterize both the practical value of BinoGen for real-world perception and its utility as a controlled framework for studying the relationship between visual experience, embodiment, and representation learning.

\subsection{Simulation-enhanced real-world perception}

We first evaluate whether synthetic visual experiences generated by BinoGen can improve perception in real-world environments. We compare three fine-tuning strategies on real-world benchmarks, including ScanNet~\citep{dai_scannet_2017} and NYUDv2~\citep{Silberman:ECCV12}. In the \textit{Zero-shot} setting, the pretrained model is evaluated directly without task-specific fine-tuning. In the \textit{Real} setting, the model is fine-tuned using only real-world training data. In the \textit{Hybrid} setting, the model is fine-tuned using both real-world data and BinoGen-generated synthetic data. The comparison between \textit{Real} and \textit{Hybrid} isolates the effect of incorporating synthetic experiences during downstream learning.

\subsubsection{Geometric perception}
\label{sec.depth}
We first evaluate depth estimation as a representative geometric perception task. Depth estimation is fundamental to embodied agents because it provides information about the three-dimensional structure of the surrounding environment. We ask whether BinoGen-generated synthetic data can improve depth estimation when only limited real-world supervision is available.

We adopt Depth Anything V2~\citep{NEURIPS2024_26cfdcd8} as the baseline model. To simulate a practical low-data adaptation scenario, we fine-tune the model based on low-rank adaptation (LoRA) using a limited number of labeled samples from ScanNet. 
We evaluate performance on ScanNet and assess cross-dataset generalization on NYUDv2 using three standard metrics: absolute relative error (AbsRel), root mean squared error (RMSE), and threshold accuracy ($\delta<1.25$).

As summarized in Table~\ref{tab:depth}, \textit{Real} fine-tuning substantially improves performance over the \textit{Zero-shot} baseline on both datasets, demonstrating the effectiveness of task-specific adaptation.
More importantly, incorporating BinoGen consistently improves all evaluation metrics. On ScanNet, \textit{Hybrid} reduces AbsRel from 0.1102 to 0.1082 and RMSE from 0.2113 to 0.2067, while slightly increasing $\delta<1.25$ from 89.34\% to 89.44\%. On NYUDv2, the gains are more pronounced: AbsRel decreases from 0.1305 to 0.1174, RMSE from 0.6510 to 0.6047, and $\delta<1.25$ increases from 81.05\% to 85.51\%.

\begin{figure}
	\begin{center}
		\includegraphics[width=\linewidth]{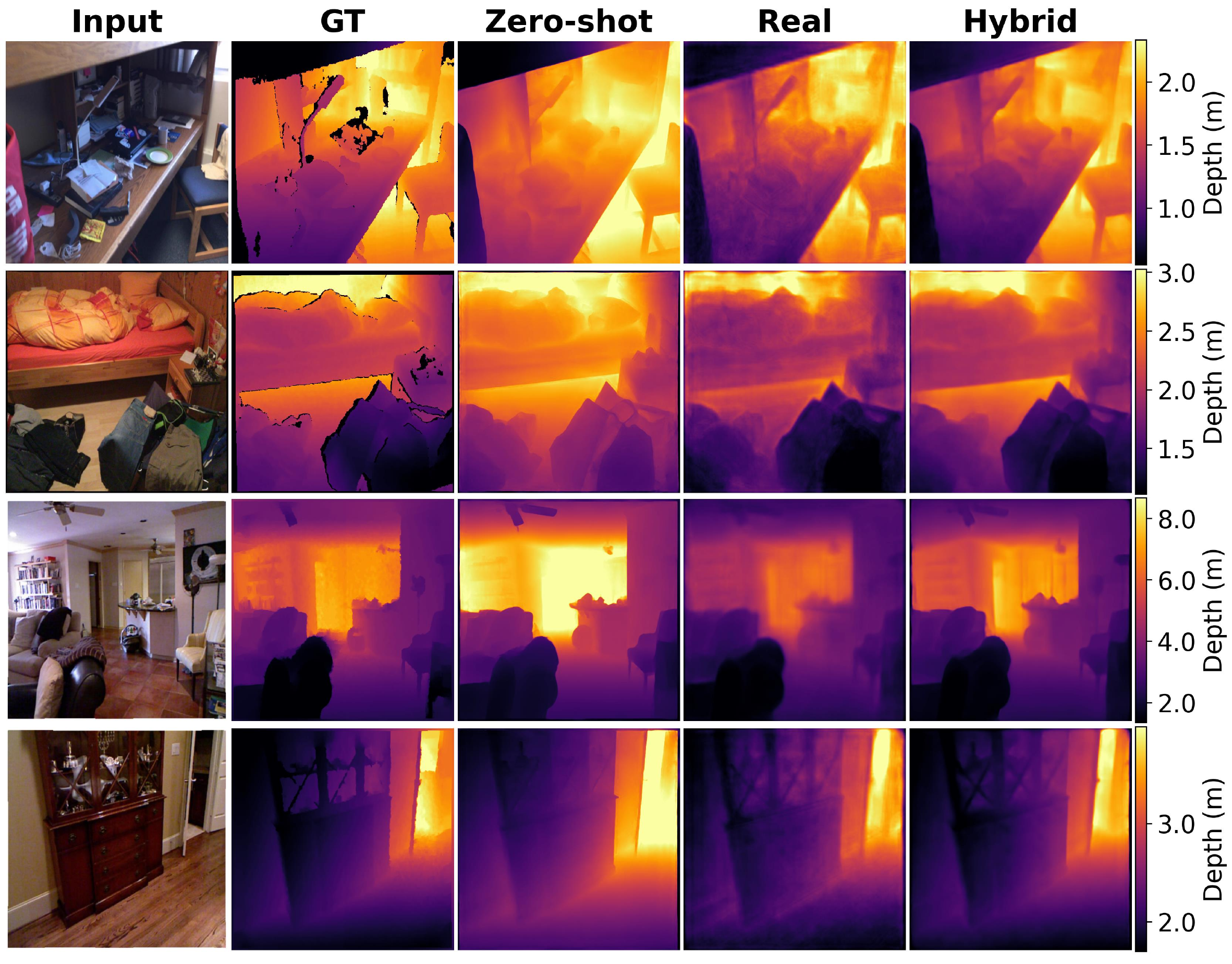}
	\end{center}
	\caption{Qualitative comparison of depth prediction on ScanNet (first two rows) and NYUDv2 (last two rows) across different models.}
	\label{fig.vis_depth}
\end{figure}

Figure~\ref{fig.vis_depth} shows that the \textit{Zero-shot} model suffers from noticeable depth-scale errors when applied directly. Fine-tuning on ScanNet largely corrects the global depth scale, but the predictions remain noisy in homogeneous regions and exhibit blurred object boundaries.
In contrast, the \textit{Hybrid} model generates smoother depth estimations with sharper structural boundaries. 
Large planar surfaces, such as walls, floors, and furniture, exhibit improved spatial consistency, while depth discontinuities around object contours are better preserved.

The consistent improvements on both ScanNet and the unseen NYUDv2 benchmark suggest that BinoGen-generated visual experiences provide complementary geometric supervision that extends beyond the specific real-world training distribution. These results demonstrate the potential of synthetic egocentric experiences to improve geometric perception under limited real-world supervision.

\begin{table*}
    \caption{Depth estimation performance on ScanNet and NYUDv2. Bold indicates the best performance.}
    \centering
    \setlength{\tabcolsep}{8pt}
    \begin{tabular}{l|ccc|ccc}
        \hline
        \multirow{2}{*}{Method} & \multicolumn{3}{c|}{ScanNet} & \multicolumn{3}{c}{NYUDv2} \\
        \cline{2-7}
        & AbsRel$\downarrow$ & RMSE$\downarrow$ & $\delta<1.25\uparrow$ & AbsRel$\downarrow$ & RMSE$\downarrow$ & $\delta<1.25\uparrow$ \\ \hline
        Zero-shot       & 0.2297 & 0.3849 & 63.50\% & 0.2008 & 0.8157 & 69.45\% \\
        Real            & 0.1102 & 0.2113 & 89.34\% & 0.1305 & 0.6510 & 81.05\% \\
        \textbf{Hybrid} & \textbf{0.1082} & \textbf{0.2067} & \textbf{89.44\%} & \textbf{0.1174} & \textbf{0.6047} & \textbf{85.51\%} \\ \hline
    \end{tabular}
    \label{tab:depth}
\end{table*}

\subsubsection{Semantic perception}
\label{sec.semantic}
We next evaluate semantic perception through object detection. Compared with geometric perception, object detection is sensitive to variations in object appearance, viewpoint, scale, illumination, and scene composition. We therefore investigate whether the diverse visual observations generated by BinoGen can provide complementary supervision for semantic perception in real-world indoor environments.

We adopt YOLOv11n~\citep{khanam2024yolov11} pretrained on COCO~\citep{lin_microsoft_2014} as the baseline detector. To ensure consistent annotation across datasets, we merge the original 45 fine-grained categories in BinoGen into 9 higher-level indoor classes and focus our evaluation on bedroom-like environments. Downstream detection and tracking are conducted on the three categories shared between BinoGen and ScanNet: \textit{bed}, \textit{stand}, and \textit{cabinet}.
We report performance on the ScanNet validation set using mean Average Precision at an IoU threshold of 0.50 (mAP${50}$) and mean Average Precision averaged over IoU thresholds from 0.50 to 0.95 (mAP$_{50:95}$). %AP$_{50}$ and AP$_{50:95}$.

\begin{figure}
	\begin{center}
		\includegraphics[width=0.7\linewidth]{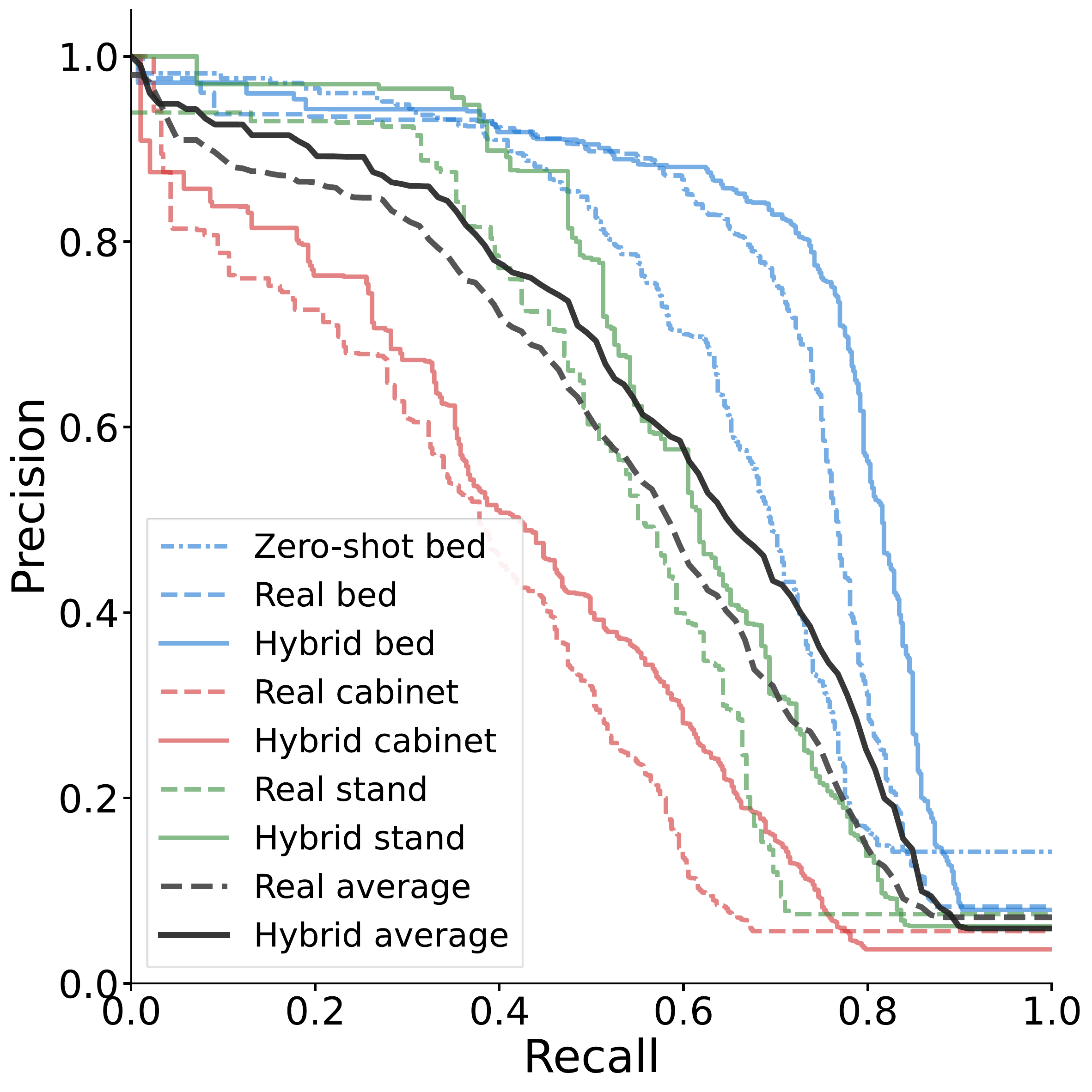}
	\end{center}
	\caption{Precision-recall curves on the ScanNet validation set for different training settings.}
	\label{fig.vis_detection}
\end{figure}

As shown in Table~\ref{tab:detection}, \textit{Real} fine-tuning substantially improves detection performance over the \textit{Zero-shot} baseline. Incorporating BinoGen-generated data further improves performance across all evaluated categories. Compared with \textit{Real}, \textit{Hybrid} increases mAP$_{50}$ from 55.6\% to 61.3\% and mAP$_{50:95}$ from 41.0\% to 45.4\%. The improvements are observed for all categories. These improvements are further illustrated by the corresponding precision-recall curves in Figure~\ref{fig.vis_detection}.

These results suggest that BinoGen provides complementary semantic supervision to limited real-world training data.
By exposing the detector to greater diversity in object appearances, viewpoints, and scene compositions, together with accurate instance-level annotations, BinoGen enables more robust semantic representation learning and improves generalization to real-world indoor scenes.

\begin{table*}
    \caption{Object detection performance on ScanNet. N/A indicates that the corresponding category is absent from the zero-shot model's original prediction label space.}
    \centering
    \setlength{\tabcolsep}{8pt}
    \begin{tabular}{l|cccc|cccc}
        \hline
        \multirow{2}{*}{Method} & \multicolumn{4}{c|}{$AP_{50}\uparrow$} & \multicolumn{4}{c}{$AP_{50:95}\uparrow$} \\
        \cline{2-9}
                  & bed  & cabinet& stand & $mAP_{50}$  & bed    &cabinet& stand & $mAP_{50:95}$ \\ \hline
        Zero-shot &67.3\% & N/A     & N/A     &  N/A    &45.0\% &  N/A   &  N/A    &  N/A    \\
        Real      &75.0\% &36.5\% &55.2\% &55.6\% &59.8\% &26.2\%&37.0\% &41.0\% \\
        \textbf{Hybrid} & \textbf{79.3\%} & \textbf{42.4\%} & \textbf{62.3\%} & \textbf{61.3\%} & 
                          \textbf{62.7\%} & \textbf{31.2\%} & \textbf{42.4\%} & \textbf{45.4\%} \\ \hline
    \end{tabular}
    \label{tab:detection}
\end{table*}

\subsubsection{Temporal perception}
\label{sec.temporal}

Beyond image-based perception, we further evaluate whether BinoGen improves temporal perception on continuous visual streams. Video understanding requires not only accurate object recognition but also consistent object association across frames. By generating temporally coherent sequences together with synchronized camera trajectories and object identities, BinoGen provides scalable supervision for learning video-based representations.

We evaluate video object tracking using CenterTrack~\citep{zhou2020tracking} as the baseline model. Performance is evaluated using mAP$_{50}$, mAP$_{50:95}$, and multiple object tracking accuracy (MOTA). 

\begin{figure}
	\begin{center}
		\includegraphics[width=0.8\linewidth]{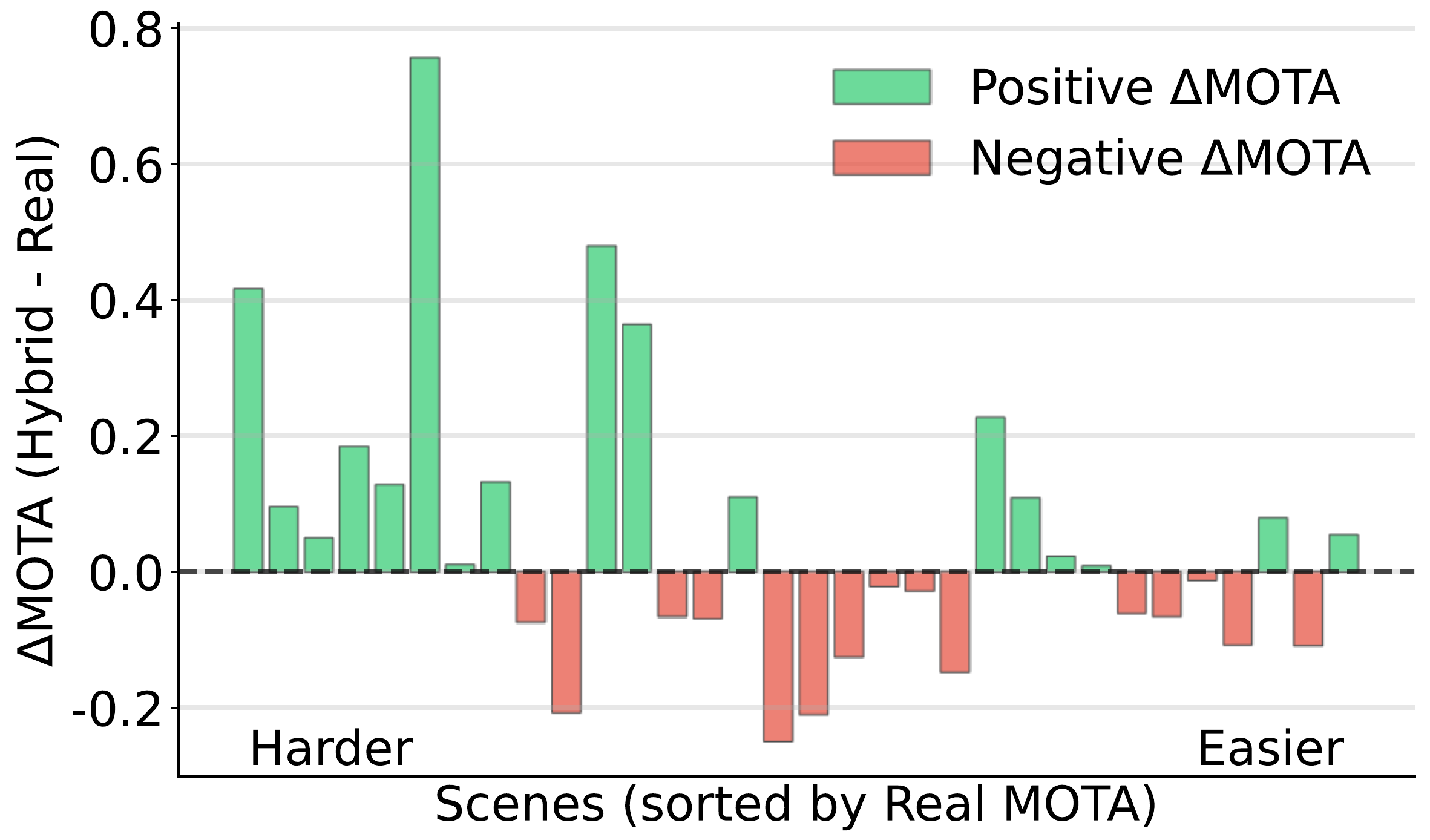}
	\end{center}
    \caption{Per-scene MOTA improvement of \textit{Hybrid} over \textit{Real}. Scenes are sorted by the MOTA of the \textit{Real} model.}
    \label{fig.vis_tracking}
\end{figure}

As shown in Table~\ref{tab:video}, incorporating BinoGen consistently improves tracking performance over real-world training alone. Compared with \textit{Real}, \textit{Hybrid} increases mAP$_{50}$ from 39.9\% to 47.6\%, mAP$_{50:95}$ from 28.8\% to 32.9\%, and MOTA from 0.43 to 0.45.
Figure~\ref{fig.vis_tracking} further examines the per-scene change in MOTA. Scenes are ordered by the MOTA achieved by the \textit{Real} model, with lower values corresponding to more challenging tracking conditions.
The improvements are more pronounced in these challenging scenes, where \textit{Hybrid} consistently achieves higher MOTA than \textit{Real}.

The gains may arise from the temporally structured visual experiences provided by BinoGen. Continuous trajectories expose objects to changing viewpoints and visibility conditions, providing multiple observations of the same objects over time. Such variation can improve object association when identities become ambiguous due to occlusion, motion, or viewpoint changes.

These results demonstrate that BinoGen complements limited real-world video data by providing scalable, temporally coherent supervision with accurate object identities. Together with the improvements observed in geometric and semantic perception, the results show that BinoGen-generated visual experiences can enhance real-world perception across spatial, semantic, and temporal dimensions.

\begin{table}
    \caption{Performance of video object tracking on ScanNet video sequences.}
    \centering
    \setlength{\tabcolsep}{8pt}
    \begin{tabular}{l|ccc}
        \hline
        Method & $mAP_{50}\uparrow$ & $mAP_{50:95}\uparrow$ & MOTA$\uparrow$ \\ \hline
        Real & 39.9\% & 28.8\% & 0.43 \\
        \textbf{Hybrid} & \textbf{47.6\%} & \textbf{32.9\%} & \textbf{0.45} \\ \hline
    \end{tabular}
    \label{tab:video}
\end{table}

\subsection{Visual perception transfer across embodiments}
\label{species}

We next investigate how differences in visual embodiment influence an agent's visual experience and the transferability of its learned perceptual representations. Even when two agents follow the same trajectory through an identical physical environment, differences in embodiment configurations (Table~\ref{tab:config}), such as body scale, camera height, and binocular geometry, can result in substantially different visual observations. Such differences introduce an embodiment-dependent visual domain gap that may limit the transfer of representations learned from one embodiment to another.

BinoGen provides a controlled framework for studying this domain gap. For each scene, we render paired visual experiences along the same trajectory using human-like and mouse-like binocular configurations. Because the physical environment and agent trajectory are identical within each pair, differences between the two visual experiences arise from the embodiment configuration rather than from changes in scene content or navigation. This paired design allows us to isolate the effect of embodiment on visual experience and directly assess the transferability of perceptual representations across embodiments.

We address three questions. First, can visual representations learned from human-viewpoint experience transfer directly to the mouse-viewpoint domain? Second, when direct transfer is insufficient, can mouse-viewpoint experience generated by BinoGen effectively adapt these representations to the target embodiment? Third, does initialization from previously learned representations provide an advantage over learning the target-embodiment task from scratch?

To answer these questions, we compare four settings. \textit{Zero-shot} directly evaluates a pretrained model without task- or domain-specific fine-tuning. \textit{Human fine-tune} adapts the pretrained model using human-viewpoint data. \textit{Mouse fine-tune} adapts the pretrained model using mouse-viewpoint data. \textit{Mouse scratch} is randomly initialized and trained using the same mouse-viewpoint training data as \textit{Mouse fine-tune}, providing a baseline for evaluating previously learned representations.

\subsubsection{Geometric perception across embodiments}
\label{sec.mouse_depth}

We first examine how embodiment differences affect the transfer of geometric perception, using depth estimation as the evaluation task. Depth perception is particularly sensitive to changes in camera height, viewpoint, and binocular geometry, which alter the relationship between observations and scene structure. It provides a direct test of whether geometric representations learned from one embodiment can generalize to another.

As summarized in Table~\ref{tab:mouse_depth}, both \textit{Zero-shot} and \textit{Human fine-tune} models perform poorly on mouse-viewpoint observations. Despite being fine-tuned on human-viewpoint observations from the same underlying environments, \textit{Human fine-tune} achieves an RMSE of 2.3104 and a threshold accuracy of only 19.5\%.
The poor direct-transfer performance reveals a substantial geometric domain gap between the human-like and mouse-like embodiments.

Target-embodiment adaptation substantially reduces this gap. Fine-tuning the human-viewpoint model with mouse-viewpoint experience (\textit{Mouse fine-tune}) decreases AbsRel from 0.0658 to 0.0400 and RMSE from 2.3104 to 0.3533, while increasing threshold accuracy from 19.5\% to 98.4\%. These improvements demonstrate that mouse-viewpoint supervision enables the model to adapt its geometric representations to the target embodiment.

Importantly, prior visual knowledge remains beneficial after target-domain adaptation. \textit{Mouse fine-tune} consistently outperforms \textit{Mouse scratch}, which is trained from random initialization using the same mouse-viewpoint data. Specifically,
it decreases AbsRel from 0.0586 to 0.0400 and RMSE from 0.4623 to 0.3533, while increasing threshold accuracy from 95.7\% to 98.4\%. 
Thus, although geometric representations learned from human-viewpoint experience do not transfer directly to the mouse-viewpoint domain, they provide a useful initialization for subsequent adaptation. These results indicate that embodiment differences substantially constrain direct perception transfer, while previously acquired visual knowledge remains valuable when combined with target-embodiment experience.

\begin{table}[pos=htb!]
    \caption{Cross-embodiment depth estimation performance on mouse-domain data.}
    \centering
    \setlength{\tabcolsep}{8pt}
    \begin{tabular}{l|ccc}
        \hline
        Method & AbsRel$\downarrow$ & RMSE$\downarrow$ & $\delta<1.25\uparrow$ \\ \hline
        Zero-shot & 0.7024 & 3.0683 & 0.9\% \\
        Human fine-tune & 0.0658 & 2.3104 & 19.5\% \\
        \textbf{Mouse fine-tune} & \textbf{0.0400} & \textbf{0.3533} & \textbf{98.4\%} \\
        Mouse scratch & 0.0586 & 0.4623 & 95.7\% \\
        \hline
    \end{tabular}
    \label{tab:mouse_depth}
\end{table}

\subsubsection{Semantic perception across embodiments}
\label{sec.cross_semantic}

Following the same evaluation protocol, we next examine whether the transfer-and-adaptation pattern observed for geometric perception extends to semantic perception.
As shown in Table~\ref{tab:mouse_detection}, direct transfer to the mouse-viewpoint domain is limited. The \textit{Zero-shot} model achieves only 5.8\% mAP${50}$ and 4.2\% mAP${50:95}$, while \textit{Human fine-tune} improves these metrics to 19.6\% and 14.2\%, respectively. However, performance remains substantially below that achieved with mouse-viewpoint supervision. Further fine-tuning on mouse-viewpoint data substantially reduces this gap, with \textit{Mouse fine-tune} reaching 84.8\% mAP${50}$ and 72.5\% mAP${50:95}$. These results indicate that semantic representations learned from human-viewpoint experience do not generalize directly to the mouse-like embodiment, but can be effectively adapted using target-embodiment experience generated by BinoGen. Importantly, prior visual knowledge remains beneficial after target-embodiment adaptation. Using the same mouse-viewpoint training data, \textit{Mouse fine-tune} consistently outperforms \textit{Mouse scratch}, which is trained from random initialization.

Taken together with the depth estimation results, these experiments reveal a consistent pattern across geometric and semantic perception. First, direct transfer from human-viewpoint to mouse-viewpoint observations is limited, even when the observations are generated from the same underlying environments and trajectories. Second, target-embodiment experience generated by BinoGen effectively bridges this embodiment-dependent perceptual gap. Third, prior visual knowledge provides additional benefits during adaptation. Thus, although embodiment-specific experience is required for effective transfer, previously acquired visual knowledge remains useful when adapting to a different visual embodiment.

\begin{table}[pos=htb!]
    \caption{Cross-embodiment object detection performance on mouse-domain data.}
    \centering
    \setlength{\tabcolsep}{8pt}
    \begin{tabular}{l|cc}
        \hline
        Method & $mAP_{50}\uparrow$ & $mAP_{50:95}\uparrow$  \\ \hline
        Zero-shot & 5.8\% & 4.2\%  \\
        Human fine-tune & 19.6\% & 14.2\%  \\
        \textbf{Mouse fine-tune} & \textbf{84.8\%} & \textbf{72.5\%}  \\
        Mouse scratch & 83.2\% & 69.5\%   \\
        \hline
    \end{tabular}
    \label{tab:mouse_detection}
\end{table}

\subsection{Joint learning across embodiments}

The preceding experiments show that visual representations learned under one embodiment do not transfer directly to another, but can provide useful initialization for target-embodiment adaptation. We next ask whether the two visual domains can instead be learned jointly, allowing a single model to develop representations that generalize across embodiments. To this end, we compare three training strategies: \textit{Human}, fine-tuning with BinoGen human-viewpoint data only; \textit{Mouse}, fine-tuning with BinoGen mouse-viewpoint data only; and \textit{Joint}, jointly fine-tuning with both human-viewpoint and mouse-viewpoint data. All models are evaluated on both visual domains using depth estimation and object detection.

As shown in Table \ref{tab:hybrid}, the \textit{Human} and \textit{Mouse} models perform best in their respective native domains but generalize poorly to the other embodiment. For object detection, the \textit{Human} model achieves 94.8\% mAP$_{50}$ on the human domain but only 19.6\% on the mouse domain, whereas the \textit{Mouse} model achieves 84.8\% on the mouse domain but only 38.9\% on the human domain. A similar asymmetry is observed for depth estimation, indicating that the embodiment gap affects both semantic and geometric perception.

In contrast, joint training substantially improves cross-embodiment generalization while preserving performance within each domain. On the human domain, the \textit{Joint} model achieves 94.4\% mAP$_{50}$, close to the 94.8\% achieved by the \textit{Human} model. On the mouse domain, it reaches 83.7\%, compared with 84.8\% for the \textit{Mouse} model. The same pattern is observed for depth estimation: \textit{Joint} remains competitive with the corresponding embodiment-specific models on both domains.

\begin{figure*}[htb!]
	\begin{center}
		\includegraphics[width=\linewidth]{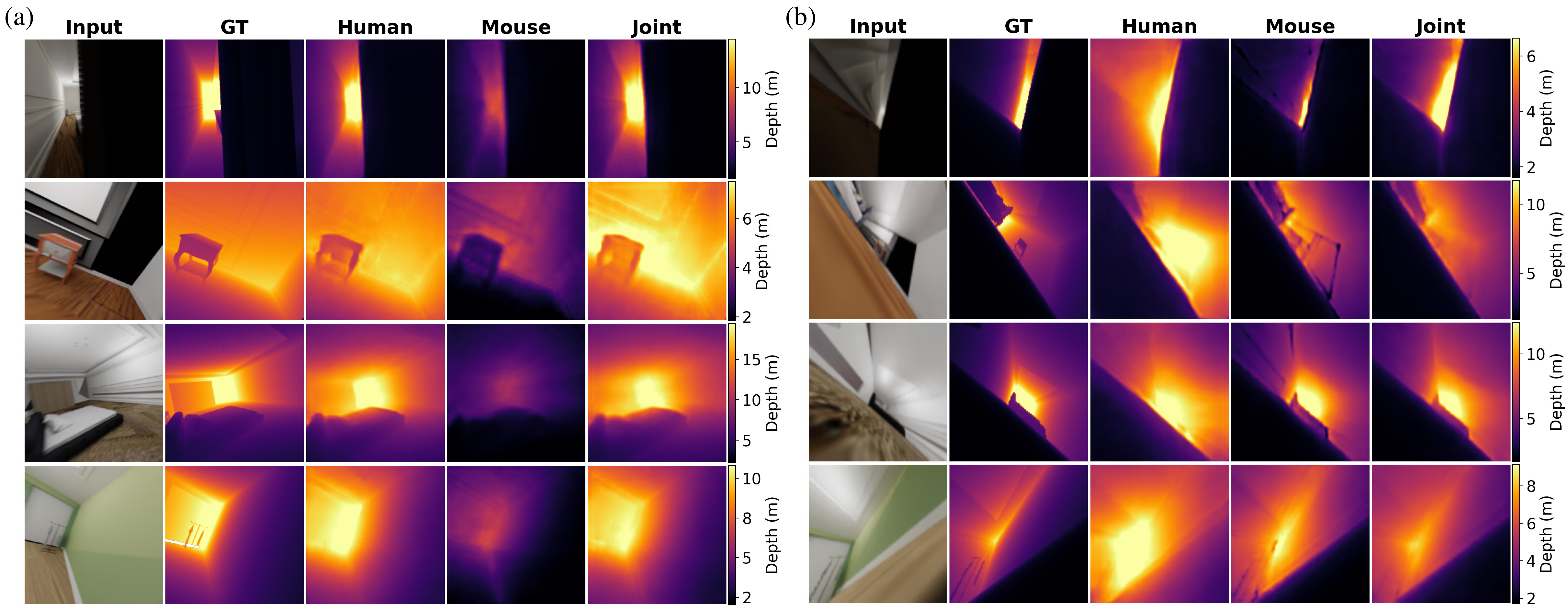}
	\end{center}
	\caption{Qualitative comparison of cross-embodiment depth prediction on paired observations from the same environments. (a) Depth predictions on human-viewpoint observations. (b) Depth predictions on corresponding mouse-viewpoint observations.}
	\label{fig.vis_mouse_depth}
\end{figure*}

Figure~\ref{fig.vis_mouse_depth} provides qualitative comparisons of depth predictions on paired human-viewpoint and mouse-viewpoint observations from the same environments. The embodiment-specific models produce accurate predictions within their native domains but show substantial degradation when applied across embodiments. In contrast, the \textit{Joint} model produces predictions that closely resemble those of the corresponding embodiment-specific models in both domains, preserving scene geometry and object boundaries. These results are consistent with the quantitative evaluation and suggest that joint training promotes geometric representations that are shared across visual embodiments.

\begin{figure}[htb!]
	\begin{center}
		\includegraphics[width=\linewidth]{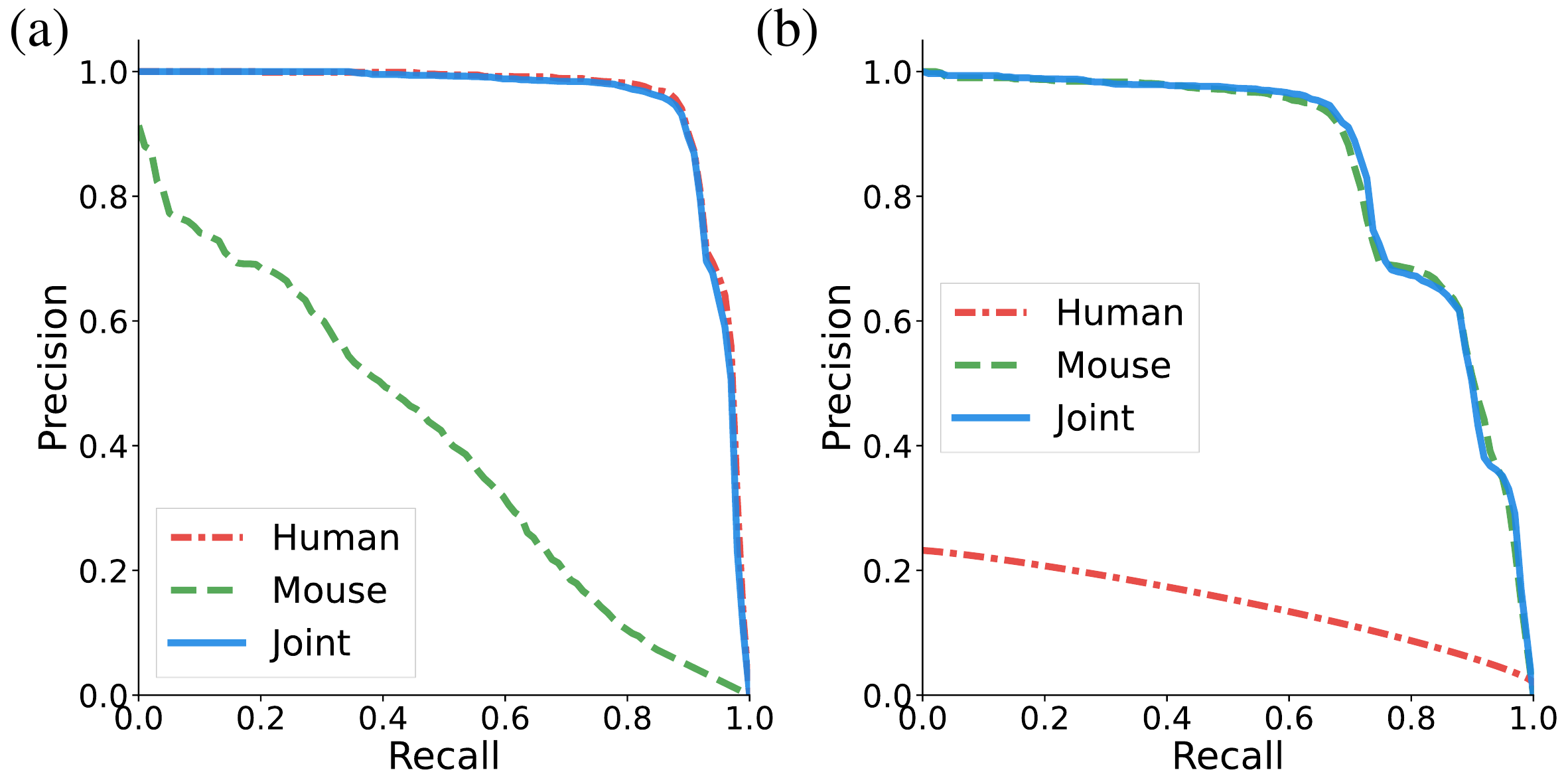}
	\end{center}
	\caption{Precision-recall curves for cross-embodiment object detection. (a) Results on human-viewpoint observations. (b) Results on the corresponding mouse-viewpoint observations.}
	\label{fig.vis_mouse_detection}
\end{figure}

Figure~\ref{fig.vis_mouse_detection} compares precision-recall curves for object detection on the two domains. Although the \textit{Human} and \textit{Mouse} models achieve the strongest performance in their respective native domains, the \textit{Joint} model remains highly competitive across both viewpoints, indicating that joint training preserves semantic recognition performance across embodiments without requiring separate detectors.

Together, these results demonstrate that paired visual experiences generated by BinoGen can support joint representation learning across substantially different visual embodiments. Rather than requiring a separate model for each embodiment, a single model trained on both human-viewpoint and mouse-viewpoint experiences can retain near-native performance in both domains. This finding indicates that embodiment-specific visual experience and embodiment-general representation learning can be achieved simultaneously when the training data provide controlled and paired variation in visual embodiment.

\begin{table*}[htb!]
    \caption{Depth estimation and object detection performance across human-viewpoint and mouse-viewpoint observations. Bold indicates the best result and underlined indicates the second-best result within each test set and task.}
    \centering
    \small
    \setlength{\tabcolsep}{8pt}
    \begin{tabular}{c|c|ccc|cc}
        \hline
        \multirow{2}{*}{Fine-tuning set} & \multirow{2}{*}{Test set} & \multicolumn{3}{c|}{Depth estimation} & \multicolumn{2}{c}{Object detection} \\
        \cline{3-7}

         &  & AbsRel$\downarrow$ & RMSE$\downarrow$ & $\delta<1.25\uparrow$ & $mAP_{50}\uparrow$ & $mAP_{50:95}\uparrow$ \\ \hline
         \multirow{2}{*}{Human}&Human &\ul{0.0420} &\textbf{0.3432} &\textbf{98.74\%} &\textbf{94.8\%} &\textbf{88.1\%} \\
                               &Mouse &0.0658 &2.3104 &19.5\% &19.6\% &14.2\% \\ \hline
         \multirow{2}{*}{Mouse}&Human &0.4777 &3.7230&2.8\% &38.9\% &26.6\% \\
                               &Mouse &\textbf{0.0400} &\textbf{0.3533} &\textbf{98.4\%} &\textbf{84.8\%} &\textbf{72.5\%} \\ \hline
        \multirow{2}{*}{Joint}&Human &\textbf{0.0410} &\ul{0.3881} &\ul{98.6\%} &\ul{94.4\%} &\ul{87.9\%}  \\
                               &Mouse &\ul{0.0447} &\ul{0.3765} &\ul{98.2\%} &\ul{83.7\%} &\ul{70.9\%}  \\ \hline
    \end{tabular}
    \label{tab:hybrid}
\end{table*}

\section{Conclusions}
\label{sec:conclusions}

In this work, we present BinoGen, an automated framework for generating embodiment-aware egocentric binocular visual experiences together with synchronized multimodal annotations. By jointly varying environmental factors, observer configurations, and visual trajectories, BinoGen enables large-scale generation of diverse binocular experiences from controllable environments. In particular, its ability to generate paired observations from different embodiments provides a systematic way to vary visual experience while controlling for the underlying scene and trajectory.

We demonstrate that BinoGen-generated visual experiences provide effective supervision for real-world perception. Across geometric, semantic, and temporal perception tasks, incorporating BinoGen-generated data consistently improves performance when combined with limited real-world supervision. These results show that synthetic visual experience can complement costly real-world data collection and annotation, providing scalable supervision for embodied visual learning.

Beyond data augmentation, BinoGen provides a controlled framework for studying how visual perception transfers across observer embodiments. Experiments using human-inspired and mouse-inspired binocular configurations reveal substantial embodiment-induced shifts in visual observations, despite identical physical environments and trajectories. Direct transfer between embodiments is limited, whereas target-embodiment experience enables effective adaptation. At the same time, previously learned visual representations remain beneficial during adaptation, and joint training on multiple embodiments enables a single model to maintain competitive performance across both visual domains. These findings highlight the complementary roles of embodiment-specific experience and shared visual representations in embodied perception.

BinoGen also provides opportunities for extending synthetic visual experience beyond the settings explored here. Future work could incorporate larger and more diverse 3D asset libraries, increase scene complexity and visual realism, and support richer agent--environment interactions and a broader range of embodied configurations. The binocular nature of the generated observations further enables applications in stereo matching, disparity estimation, binocular tracking, and multimodal embodied learning.

More broadly, BinoGen shifts the focus from generating synthetic images to generating controllable visual experiences. By jointly modeling the environment, observer, and trajectory, it provides a flexible framework for investigating how visual experience shapes perception across different embodied agents. We hope that BinoGen will serve as a useful resource for research in embodied perception, multimodal learning, stereo vision, and embodiment-aware artificial intelligence.

\section*{Appendix}
\appendix

\section{Experimental setup}
This appendix provides additional details of the downstream evaluation, including training protocols, dataset construction, and evaluation metrics.

\subsection{Common dataset setup}
\label{sec.a1}

For simulation-enhanced real-world perception experiments, real-world data are sourced from the apartment and hotel subsets of ScanNet, whereas synthetic data are generated from the bedroom subset of BinoGen. Although the synthetic and real-world datasets are not perfectly matched, both represent indoor residential environments, reducing substantial scene-domain differences while preserving a realistic evaluation setting.

For cross-species experiments, both human-viewpoint and mouse-viewpoint observations are rendered from the same set of BinoGen bedroom environments. Differences in visual observations are introduced through embodiment-specific camera configurations.

To prevent data leakage, all training and validation splits are performed at the scene level rather than at the image level. Specifically, images or video frames from the same environment are assigned exclusively to either the training or validation split. Consequently, no scene is shared between the training and evaluation sets, ensuring that performance reflects generalization to unseen environments rather than memorization of scene-specific layouts or object configurations.

For image-based perception tasks, we intentionally employ a limited amount of real-world supervision. Rather than using the full set of available annotations, we construct relatively small, task-specific training sets to mimic practical embodied perception scenarios in which large-scale labeled data are difficult to acquire. This data-constrained setting enables us to directly assess the contribution of BinoGen-generated synthetic experiences.  Improvements achieved by incorporating synthetic data under these conditions demonstrate that BinoGen can provide complementary supervision when real-world observations are limited.

\subsection{Depth estimation}

\subsubsection{Model and training protocol}

\paragraph{Model.}  
We use Depth Anything V2 with a ViT-S backbone as the base model for depth estimation.
To enable parameter-efficient adaptation, we fine-tune the model using low-rank adaptation (LoRA). Specifically, LoRA modules are inserted into the query, key, and value projection layers of every self-attention block, while the original backbone parameters remain frozen.
We apply LoRA with rank $r=8$, scaling factor $\alpha=16$, and a dropout rate of $0.05$. The feature pyramid network (FPN) neck and prediction head are kept fully trainable. In total, 3.0M parameters are optimized during training, corresponding to approximately 12\% of the full model, while 22.1M parameters are kept frozen.

\paragraph{Loss function.}
Metric depth prediction is supervised using the scale-invariant logarithmic (SILog) loss. Let $d_i$ and $d_i^{*}$ denote the predicted and ground-truth depths, respectively, and define
\begin{equation} 
g_i=\log d_i-\log d_i^{*}.
\end{equation}
The SILog loss is computed as
\begin{equation} 
L_{\mathrm{SILog}} = \frac{1}{N}\sum_i g_i^2 - 0.5 \left( \frac{1}{N}\sum_i g_i \right)^2,
\end{equation}
where $N$ denotes the number of pixels with valid depth annotations. The loss is evaluated only over pixels with valid depth annotations. 

\paragraph{Training.}
All trainable parameters are jointly optimized using AdamW. Unless otherwise specified, all experiments are trained for 15 epochs using a batch size of 8, a fixed learning rate of $5\times10^{-5}$, a weight decay of $1\times10^{-4}$, and an input resolution of $518\times518$ pixels. To improve training stability, gradient clipping with a maximum norm of 1.0 is applied throughout training. For data augmentation, horizontal flipping is applied with a probability of 0.5, with the RGB image and corresponding depth map flipped jointly.

\subsubsection{Evaluation}
Following standard monocular depth estimation protocols, we report the absolute relative error (AbsRel), root mean square error (RMSE), and threshold accuracy ($\delta < 1.25$). AbsRel is defined as
\begin{equation}
\mathrm{AbsRel}=\frac{1}{N}\sum_i\frac{|d_i-d_i^{*}|}{d_i^{*}}\end{equation}
and RMSE is computed as
\begin{equation}
\mathrm{RMSE}=\sqrt{\frac{1}{N}\sum_i(d_i-d_i^{*})^2}\end{equation}
Threshold accuracy measures the fraction of valid depth pixels for which
\begin{equation}
\max\left(\frac{d_i}{d_i^{*}},\frac{d_i^{*}}{d_i}\right)<1.25\end{equation}

\subsubsection{Datasets}

\paragraph{Simulation-enhanced real-world perception.}

For the experiments in Sec.\ref{sec.depth}, the \textit{Real} setting uses ScanNet data only, whereas the \textit{Hybrid} setting combines ScanNet with BinoGen-generated synthetic samples.
To establish a data-constrained adaptation setting and reduce temporal redundancy, frames from the ScanNet training split are sampled every 500 frames, yielding 1,231 training images from 274 indoor scenes. All images are resized to $518\times518$ pixels.
For synthetic training data, we randomly sample 2,000 images from BinoGen, covering 445 indoor environments. To reduce the depth-distribution mismatch between BinoGen and ScanNet, all synthetic depth maps are linearly rescaled such that the average scene depth is 2.5~m.
In the \textit{Hybrid} setting, the ScanNet and BinoGen samples are combined into a single training set containing 3,231 images. All samples are shuffled and sampled uniformly during training.

Evaluation is performed on both the ScanNet validation split and the official NYUDv2 validation split. The ScanNet validation set contains 202 images from 39 scenes, while NYUDv2 contains 249 images.

\paragraph{Cross-embodiment visual adaptation.}
For the experiments in Sec.~\ref{sec.mouse_depth}, we construct paired human-viewpoint and mouse-viewpoint datasets from the same set of synthetic environments.
Specifically, 500 indoor scenes are rendered with both human-like and mouse-like configurations. To isolate the effect of embodiment, mouse trajectories are obtained by transferring the corresponding human trajectories to the mouse agent while adopting mouse-like camera configurations. Thus, the underlying environments and trajectories are held constant while the visual observations differ according to the embodiment.

To reduce temporal redundancy, 20 frames are sampled from each scene, and both left-eye and right-eye views are rendered for every sampled frame, which results in 10000 human-viewpoint images and 10000 mouse-viewpoint images in total.
Training and validation splits are performed at the scene level to prevent environment-level data leakage. We allocate 450 scenes (90\%) to the training set and 50 scenes (10\%) to the validation set. The same scene split is used for both the human-viewpoint and mouse-viewpoint datasets, ensuring that evaluation measures adaptation across embodied visual experiences rather than memorization of scene-specific content.

\subsection{Object detection}

\subsubsection{Model and training protocol}

\paragraph{Model.}
We use YOLOv11n initialized with COCO-pretrained weights for object detection. The model contains approximately 2.6M parameters, and the detection head is configured to predict three object categories: \textit{bed}, \textit{cabinet}, and \textit{stand}. During inference, up to 300 detections are retained per image, with non-maximum suppression (NMS) applied using an IoU threshold of 0.7.

\paragraph{Loss function.}
YOLOv11n optimizes a multi-task objective comprising bounding-box regression, object classification, and localization quality estimation. Bounding-box regression is supervised using Complete IoU (CIoU) loss, which jointly considers the overlap between predicted and ground-truth boxes, their center distance, and aspect-ratio consistency. Object classification is optimized using binary cross-entropy (BCE) loss. Distribution Focal Loss (DFL) models bounding-box coordinates as discrete probability distributions, improving localization precision.
The final loss is given by
\begin{equation}
L=\lambda_{\mathrm{box}}L_{\mathrm{CIoU}}
+\lambda_{\mathrm{cls}}L_{\mathrm{BCE}}
+\lambda_{\mathrm{DFL}}L_{\mathrm{DFL}},
\end{equation}
where the loss weights are
$\lambda_{\mathrm{box}}=7.5$,
$\lambda_{\mathrm{cls}}=0.5$,
and
$\lambda_{\mathrm{DFL}}=1.5$.

\paragraph{Training.}
Models are trained for 75 epochs with a batch size of 16. Optimization is performed using AdamW with an initial learning rate of $5\times10^{-3}$ and weight decay of $5\times10^{-4}$. A three-epoch warmup is followed by linear learning-rate decay. We apply standard YOLO data augmentations, including color jittering, random scaling, translation, horizontal flipping, mosaic augmentation, RandAugment~\citep{NEURIPS2020_d85b63ef}, and random erasing. Mosaic augmentation is disabled during the final 10 epochs to facilitate convergence. Exponential moving average (EMA) weights are used for evaluation.

\subsubsection{Evaluation}
\label{sec.det_eval}
Following the COCO evaluation protocol, we report mean average precision at an IoU threshold of 0.5 (AP$_{50}$) and mean average precision averaged over IoU thresholds from 0.5 to 0.95 with a step size of 0.05 (AP$_{50:95}$).
AP$_{50}$ measures the average precision when a predicted bounding box is considered correct if its IoU with the ground-truth bounding box exceeds 0.5. AP$_{50:95}$ averages the AP scores computed at ten IoU thresholds $\{0.50,0.55,\ldots,0.95\}$ and provides a more stringent assessment of localization accuracy. Per-class AP scores are additionally reported for the evaluated object categories.

\subsubsection{Datasets}

\begin{figure}[htb!]
	\begin{center}
		\includegraphics[width=1\linewidth]{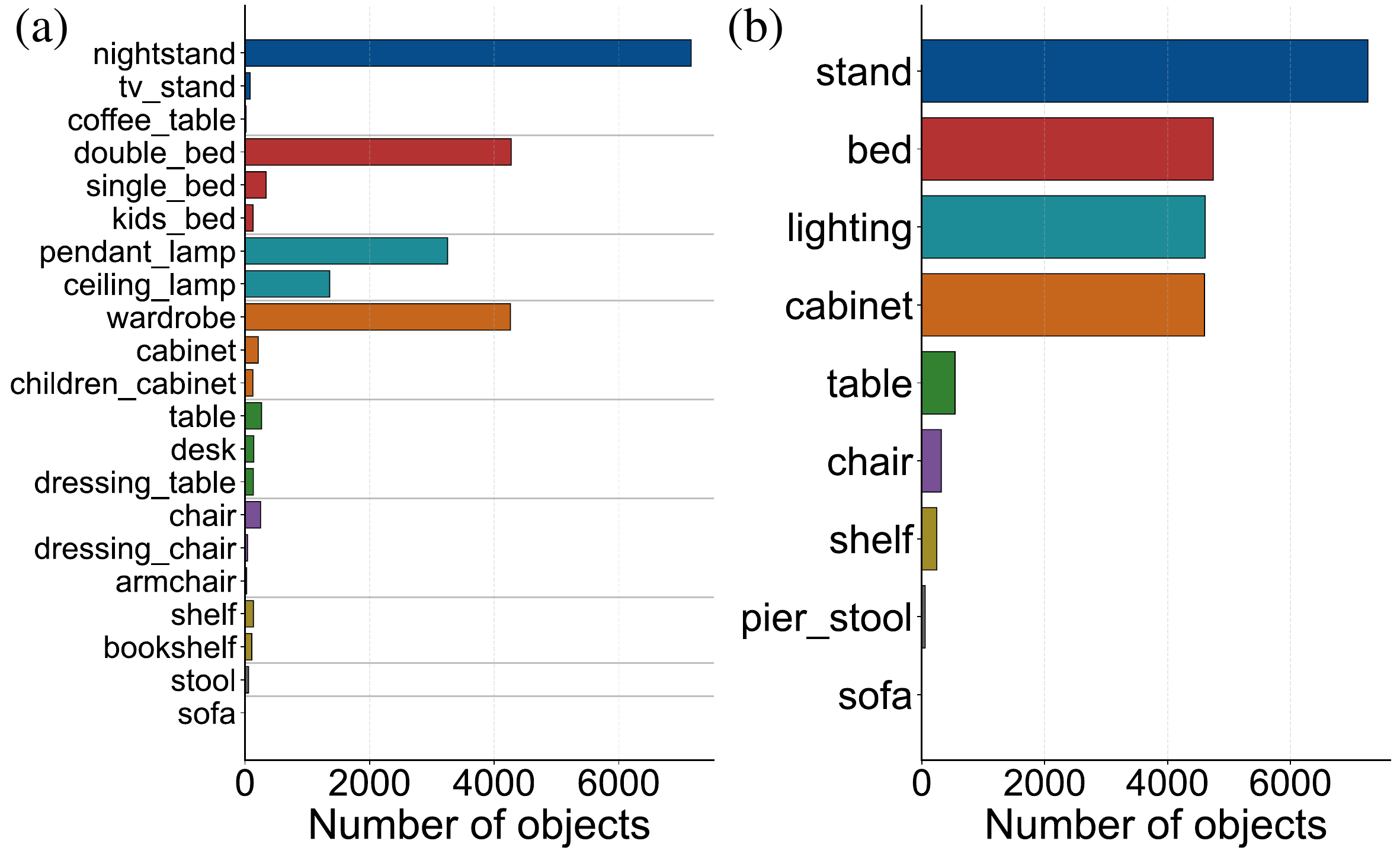}
	\end{center}
	\caption{Distribution of fine-grained object categories inherited from 3D-FRONT and the merged coarse categories used for object detection.}
	\label{fig.9cls}
\end{figure}

\paragraph{Category mapping}
BinoGen inherits 45 fine-grained object categories defined in 3D-FRONT~\citep{fu20213d}. As shown in Fig.~\ref{fig.9cls}a, 21 of these object categories are presented in the generated bedroom subset. For object detection, these fine-grained categories are merged into nine coarse semantic categories (Fig.~\ref{fig.9cls}b). This mapping serves two purposes: first, it improves category balance in the generated dataset; second, it aligns the semantic granularity with the ScanNet benchmark~\citep{dai_scannet_2017} to enable consistent sim-to-real evaluation. The complete mapping is summarized in Table~\ref{tab.9cls}.

Although nine coarse categories are defined for synthetic data generation, only categories shared between BinoGen and the ScanNet annotations are used in downstream evaluation. Consequently, the real-world object detection experiments in Sec.~\ref{sec.semantic} focus on three categories: \textit{bed}, \textit{cabinet}, and \textit{stand}.
Bounding-box annotations for synthetic images are generated automatically from the object instance segmentation maps rendered by BinoGen and converted to YOLO format.

\begin{table}[htb!]
    \centering
    \small
    \caption{Mapping from fine-grained object categories to the nine coarse categories used for object detection.}
    \begin{tabularx}{.45\textwidth}{cX}
        \midrule
        \textbf{Coarse categories} &
        \textbf{Fine-grained categories} \\
        \midrule
        bed &
        double bed, kids bed, single bed \\
        cabinet &
        cabinet, children cabinet, wardrobe \\
        chair &
        armchair, chair, dressing chair \\
        lighting &
        ceiling lamp, pendant lamp \\
        shelf &
        bookshelf, shelf \\
        stand &
        nightstand, TV stand \\
        table &
        coffee table, desk, dressing table, table \\
        stool & pier stool \\
        sofa & sofa\\
        \midrule
    \end{tabularx}
    \label{tab.9cls}
\end{table}

\paragraph{Simulation-enhanced real-world perception.}
The \textit{Real} setting uses a ScanNet-only training set containing 4,179 images sampled from 273 ScanNet scenes.
The validation set contains 740 images from 39 independent ScanNet scenes with no scene overlap between the training and validation sets.
The \textit{Hybrid} setting augments the ScanNet training set with 8,425 BinoGen-generated synthetic images from 434 bedroom environments. The resulting Hybrid training set contains 12,604 images, including 4,179 real images (33.2\%) and 8,425 synthetic images (66.8\%), covering 707 unique environments in total.
For both settings, images are resized to $640\times640$ pixels. The same ScanNet validation split containing 740 images is used for evaluation to ensure a fair comparison between the Real and Hybrid settings. Real and synthetic samples are merged into a single training set and sampled uniformly during optimization.

\paragraph{Cross-embodiment visual adaptation.}
For the cross-embodiment object detection experiments in Sec.~\ref{sec.cross_semantic}, we use the same paired human-viewpoint and mouse-viewpoint datasets described in Sec.~\ref{sec.a1}. Human-viewpoint and mouse-viewpoint annotations are generated automatically from the corresponding synthetic environments. The resulting dataset contains 10,000 human-viewpoint and 10,000 mouse-viewpoint images with automatically generated object annotations. Scene-level train/validation splits are shared across both domains, ensuring that embodied visual differences remain the primary source of domain variation. All images are resized to $512\times512$ pixels.

\subsection{Video object tracking}
\subsubsection{Model and training protocol}
\paragraph{Model}
Video object tracking experiments are conducted using CenterTrack with a DLA-34 backbone and deformable convolutions (DCNv2~\citep{Zhu_2019_CVPR}). The model contains approximately 18.7M parameters and is initialized with COCO tracking-pretrained weights. It is then fine-tuned to detect and track the three target categories (\textit{bed}, \textit{cabinet}, and \textit{stand}). Following the original CenterTrack formulation, the model takes the current RGB frame, the previous RGB frame, and the previous-frame detection heatmap as inputs, and jointly predicts object detections and tracking associations.

\paragraph{Loss function}
The overall training objective follows the original CenterTrack implementation and comprises five components:
\begin{equation}
\mathcal{L} =\mathcal{L}_{hm}+\mathcal{L}_{reg}+\lambda_{wh}\mathcal{L}_{wh}+\lambda_{ltrb}\mathcal{L}_{ltrb}+\mathcal{L}_{track},
\end{equation}
where $\mathcal{L}_{hm}$ denotes the heatmap loss for object center localization, 
$\mathcal{L}_{reg}$ is the center offset regression loss that compensates for feature map downsampling quantization errors, 
$\mathcal{L}_{wh}$ corresponds to the bounding box width and height regression loss, 
$\mathcal{L}_{ltrb}$ supervises the regression of the distances from an object center to the left, top, right, and bottom boundaries of its bounding box,
and $\mathcal{L}_{track}$ represents the tracking regression loss for inter-frame object displacement. 
Following the original implementation, the loss weights are set to $\lambda_{wh}=0.1$ and $\lambda_{ltrb}=0.1$.

\paragraph{Training}
All model parameters are fine-tuned for 70 epochs using Adam with a batch size of 32 and an initial learning rate of $5\times10^{-4}$. A multi-step learning-rate schedule is adopted, with learning-rate decay applied at epochs 45 and 60. Input frames are resized to $512\times512$ pixels.
Following the original training protocols of CornerNet\cite{Law_2018_ECCV} and CenterTrack, we apply appearance augmentations including brightness, contrast, saturation, and PCA-based lighting perturbations. The same augmentation parameters are applied to both frames in each training pair to preserve temporal consistency.

To improve robustness to imperfect detections, we introduce temporal noise during training. Specifically, heatmap disturbance is applied with a magnitude of 0.05, missed detections are simulated with a probability of 0.4, and false-positive detections are introduced with a probability of 0.1. Frame pairs are sampled with a maximum temporal distance of three frames.

\subsubsection{Evaluation}

Video tracking performance is evaluated using both detection and tracking metrics.
Detection accuracy is measured using AP$_{50}$ and AP$_{50:95}$ following the COCO evaluation protocol described in Sec.~\ref{sec.det_eval}.
Tracking performance is measured using Multiple Object Tracking Accuracy (MOTA),
\begin{equation}
\mathrm{MOTA}=1-\frac{\mathrm{FN}+\mathrm{FP}+\mathrm{IDSW}}{\mathrm{GT}},
\end{equation}
where FN, FP, and IDSW denote the numbers of false negatives, false positives, and identity switches, respectively, and GT is the number of ground-truth objects.

\subsubsection{Datasets}
\paragraph{Simulation-enhanced real-world perception.}
For the experiments in Sec.~\ref{sec.temporal}, we construct a three-category video tracking benchmark from ScanNet. We retain sequences containing at least one instance of the target categories \textit{bed}, \textit{cabinet}, or \textit{stand}, and use annotations for these same categories as in the image-based detection experiments.
The ScanNet training split contains 248 video sequences, comprising 33,325 frames and 55,898 object annotations.
The validation split contains 36 sequences, 4,790 frames, and 9,614 object annotations.

For the \textit{Hybrid} setting, we augment the ScanNet training data with 1,551 BinoGen-generated video sequences, comprising 121,314 frames and 198,945 automatically generated object annotations.
Synthetic bounding boxes and object identities are obtained directly from the rendering engine.
The resulting Hybrid training set contains 1,799 video sequences, 154,639 frames, and 254,843 object annotations.
No balancing or re-weighting is applied; ScanNet and BinoGen sequences are concatenated and sampled uniformly during training.

\section{Code and dataset release}

We will release the complete BinoGen framework upon publication, including the data generation pipeline and the complete binocular video dataset with annotations.

\printcredits

\section*{Declaration of generative AI and AI-assisted technologies in the manuscript preparation process}
During the preparation of this work, the authors used ChatGPT and DeepSeek in order to improve the language of this paper. After using this tool/service, the authors reviewed and edited the content as needed and take full responsibility for the content of the published article.

\bibliographystyle{cas-model2-names}


\bibliography{main}

@String(CVPR= {IEEE Conf. Comput. Vis. Pattern Recog.})

@String(ICCV= {Int. Conf. Comput. Vis.})

@String(ECCV= {Eur. Conf. Comput. Vis.})

@String(ICPR = {Int. Conf. Pattern Recog.})

@String(CVPR  = {CVPR})

@String(ICCV  = {ICCV})

@String(ECCV  = {ECCV})

@String(ICPR  = {ICPR})

@inproceedings{dai_scannet_2017,
	author = {Dai, Angela and Chang, Angel X. and Savva, Manolis and Halber, Maciej and Funkhouser, Thomas and Niessner, Matthias},
	title = {ScanNet: Richly-Annotated 3D Reconstructions of Indoor Scenes},
	booktitle = {Proceedings of the IEEE/CVF Conference on Computer Vision and Pattern Recognition (CVPR)},
	month = {July},
    pages={2432-2443},
	year = {2017}
}

@inproceedings{armeni_3d_2016,
	author = {Armeni, Iro and Sener, Ozan and Zamir, Amir R. and Jiang, Helen and Brilakis, Ioannis and Fischer, Martin and Savarese, Silvio},
	title = {3D Semantic Parsing of Large-Scale Indoor Spaces},
	booktitle = {Proceedings of the IEEE/CVF Conference on Computer Vision and Pattern Recognition (CVPR)},
	month = {June},
    pages={1534-1543},
	year = {2016}
}

@inproceedings{grauman_ego4d_2022,
	author    = {Grauman, Kristen and Westbury, Andrew and Byrne, Eugene and Chavis, Zachary and Furnari, Antonino and Girdhar, Rohit and Hamburger, Jackson and Jiang, Hao and Liu, Miao and Liu, Xingyu and Martin, Miguel and Nagarajan, Tushar and Radosavovic, Ilija and Ramakrishnan, Santhosh Kumar and Ryan, Fiona and Sharma, Jayant and Wray, Michael and Xu, Mengmeng and Xu, Eric Zhongcong and Zhao, Chen and Bansal, Siddhant and Batra, Dhruv and Cartillier, Vincent and Crane, Sean and Do, Tien and Doulaty, Morrie and Erapalli, Akshay and Feichtenhofer, Christoph and Fragomeni, Adriano and Fu, Qichen and Gebreselasie, Abrham and Gonz\'alez, Cristina and Hillis, James and Huang, Xuhua and Huang, Yifei and Jia, Wenqi and Khoo, Weslie and Kol\'a\v{r}, J\'achym and Kottur, Satwik and Kumar, Anurag and Landini, Federico and Li, Chao and Li, Yanghao and Li, Zhenqiang and Mangalam, Karttikeya and Modhugu, Raghava and Munro, Jonathan and Murrell, Tullie and Nishiyasu, Takumi and Price, Will and Ruiz, Paola and Ramazanova, Merey and Sari, Leda and Somasundaram, Kiran and Southerland, Audrey and Sugano, Yusuke and Tao, Ruijie and Vo, Minh and Wang, Yuchen and Wu, Xindi and Yagi, Takuma and Zhao, Ziwei and Zhu, Yunyi and Arbel\'aez, Pablo and Crandall, David and Damen, Dima and Farinella, Giovanni Maria and Fuegen, Christian and Ghanem, Bernard and Ithapu, Vamsi Krishna and Jawahar, C. V. and Joo, Hanbyul and Kitani, Kris and Li, Haizhou and Newcombe, Richard and Oliva, Aude and Park, Hyun Soo and Rehg, James M. and Sato, Yoichi and Shi, Jianbo and Shou, Mike Zheng and Torralba, Antonio and Torresani, Lorenzo and Yan, Mingfei and Malik, Jitendra},
    title     = {Ego4D: Around the World in 3,000 Hours of Egocentric Video},
    booktitle = {Proceedings of the IEEE/CVF Conference on Computer Vision and Pattern Recognition (CVPR)},
    month     = {June},
    year      = {2022},
    pages     = {18995-19012}
}

@inproceedings{mccormac_scenenet_2017,
	author = {McCormac, John and Handa, Ankur and Leutenegger, Stefan and Davison, Andrew J.},
	title = {SceneNet RGB-D: Can 5M Synthetic Images Beat Generic ImageNet Pre-Training on Indoor Segmentation?},
	booktitle = {Proceedings of the IEEE/CVF International Conference on Computer Vision (ICCV)},
	month = {Oct},
    pages={2697-2706},
	year = {2017}
}

@inproceedings{paschalidou_atiss_2021,
	author = {Paschalidou, Despoina and Kar, Amlan and Shugrina, Maria and Kreis, Karsten and Geiger, Andreas and Fidler, Sanja},
	booktitle = {Advances in Neural Information Processing Systems},
	pages = {12013--12026},
	title = {ATISS: Autoregressive Transformers for Indoor Scene Synthesis},
	volume = {34},
	year = {2021}
}

@inproceedings{chang_matterport3d_2017,
	title = {Matterport3D: Learning from RGB-D Data in Indoor Environments},
	shorttitle = {{Matterport3D}},
	booktitle = {{International} {Conference} on {3D} {Vision} ({3DV})},
	author = {Chang, Angel and Dai, Angela and Funkhouser, Thomas and Halber, Maciej and Niebner, Matthias and Savva, Manolis and Song, Shuran and Zeng, Andy and Zhang, Yinda},
	month = oct,
	year = {2017},
	pages = {667--676},
}

@inproceedings{damen_scaling_2018,
	author = {Damen, Dima and Doughty, Hazel and Farinella, Giovanni Maria and Fidler, Sanja and Furnari, Antonino and Kazakos, Evangelos and Moltisanti, Davide and Munro, Jonathan and Perrett, Toby and Price, Will and Wray, Michael},
	title = {Scaling Egocentric Vision: The EPIC-KITCHENS Dataset},
	booktitle = {Proceedings of the European Conference on Computer Vision (ECCV)},
	month = {September},
    pages={720-736},
	year = {2018}
}

@inproceedings{sigurdsson_actor_2018,
	author = {Sigurdsson, Gunnar A. and Gupta, Abhinav and Schmid, Cordelia and Farhadi, Ali and Alahari, Karteek},
	title = {Actor and Observer: Joint Modeling of First and Third-Person Videos},
	booktitle = {Proceedings of the IEEE/CVF Conference on Computer Vision and Pattern Recognition (CVPR)},
	month = {June},
    pages={7396-7404},
	year = {2018}
}

@article{chang_shapenet_2015,
	title = {{ShapeNet}: {An} {Information}-{Rich} {3D} {Model} {Repository}},
	shorttitle = {{ShapeNet}},
	urldate = {2024-05-21},
	author = {Chang, Angel X. and Funkhouser, Thomas and Guibas, Leonidas and Hanrahan, Pat and Huang, Qixing and Li, Zimo and Savarese, Silvio and Savva, Manolis and Song, Shuran and Su, Hao and Xiao, Jianxiong and Yi, Li and Yu, Fisher},
	month = dec,
	year = {2015},
	journal={arXiv preprint arXiv:1512.03012},
}

@inproceedings{wu_3d_2015,
	author = {Wu, Zhirong and Song, Shuran and Khosla, Aditya and Yu, Fisher and Zhang, Linguang and Tang, Xiaoou and Xiao, Jianxiong},
	title = {3D ShapeNets: A Deep Representation for Volumetric Shapes},
	booktitle = {Proceedings of the IEEE Conference on Computer Vision and Pattern Recognition (CVPR)},
	month = {June},
    pages={1912-1920},
	year = {2015}
}

@inproceedings{fu20213d,
  title={3d-front: 3d furnished rooms with layouts and semantics},
  author={Fu, Huan and Cai, Bowen and Gao, Lin and Zhang, Ling-Xiao and Wang, Jiaming and Li, Cao and Zeng, Qixun and Sun, Chengyue and Jia, Rongfei and Zhao, Binqiang and others},
  booktitle={Proceedings of the IEEE/CVF International Conference on Computer Vision (ICCV)},
  pages={10933--10942},
  year={2021}
}

@inproceedings{tang2024diffuscene,
  title={Diffuscene: Denoising diffusion models for generative indoor scene synthesis},
  author={Tang, Jiapeng and Nie, Yinyu and Markhasin, Lev and Dai, Angela and Thies, Justus and Nie{\ss}ner, Matthias},
  booktitle={Proceedings of the IEEE/CVF Conference on Computer Vision and Pattern Recognition (CVPR)},
  pages={20507-20518},
  year={2024}
}

@InProceedings{Wei_2023_CVPR,
    author    = {Wei, Qiuhong Anna and Ding, Sijie and Park, Jeong Joon and Sajnani, Rahul and Poulenard, Adrien and Sridhar, Srinath and Guibas, Leonidas},
    title     = {LEGO-Net: Learning Regular Rearrangements of Objects in Rooms},
    booktitle = {Proceedings of the IEEE/CVF Conference on Computer Vision and Pattern Recognition (CVPR)},
    month     = {June},
    year      = {2023},
    pages     = {19037-19047}
}

@InProceedings{Huang_2023_CVPR,
    author    = {Huang, Siyuan and Wang, Zan and Li, Puhao and Jia, Baoxiong and Liu, Tengyu and Zhu, Yixin and Liang, Wei and Zhu, Song-Chun},
    title     = {Diffusion-Based Generation, Optimization, and Planning in 3D Scenes},
    booktitle = {Proceedings of the IEEE/CVF Conference on Computer Vision and Pattern Recognition (CVPR)},
    month     = {June},
    year      = {2023},
    pages     = {16750-16761}
}

@inproceedings{li2024egogen, 
 title={{EgoGen: An Egocentric Synthetic Data Generator}}, 
 author={Li, Gen and Zhao, Kaifeng and Zhang, Siwei and Lyu, Xiaozhong and Dusmanu, Mihai and Zhang, Yan and Pollefeys, Marc and Tang, Siyu}, 
 booktitle={Proceedings of the IEEE/CVF Conference on Computer Vision and Pattern Recognition (CVPR)}, 
 pages={14497-14509},
 year={2024} 
}

@INPROCEEDINGS{6247820,
  author={Lee, Yong Jae and Ghosh, Joydeep and Grauman, Kristen},
  booktitle={Proceedings of the IEEE/CVF Conference on Computer Vision and Pattern Recognition (CVPR)}, 
  title={Discovering important people and objects for egocentric video summarization}, 
  year={2012},
  volume={},
  number={},
  pages={1346-1353},
}

@INPROCEEDINGS{6977451,
  author={Iwashita, Yumi and Takamine, Asamichi and Kurazume, Ryo and Ryoo, M.S.},
  booktitle={International Conference on Pattern Recognition (ICPR)}, 
  title={First-Person Animal Activity Recognition from Egocentric Videos}, 
  year={2014},
  volume={},
  number={},
  pages={4310-4315},
}

@InProceedings{Greff_2022_CVPR,
    author    = {Greff, Klaus and Belletti, Francois and Beyer, Lucas and Doersch, Carl and Du, Yilun and Duckworth, Daniel and Fleet, David J. and Gnanapragasam, Dan and Golemo, Florian and Herrmann, Charles and Kipf, Thomas and Kundu, Abhijit and Lagun, Dmitry and Laradji, Issam and Liu, Hsueh-Ti (Derek) and Meyer, Henning and Miao, Yishu and Nowrouzezahrai, Derek and Oztireli, Cengiz and Pot, Etienne and Radwan, Noha and Rebain, Daniel and Sabour, Sara and Sajjadi, Mehdi S. M. and Sela, Matan and Sitzmann, Vincent and Stone, Austin and Sun, Deqing and Vora, Suhani and Wang, Ziyu and Wu, Tianhao and Yi, Kwang Moo and Zhong, Fangcheng and Tagliasacchi, Andrea},
    title     = {Kubric: A Scalable Dataset Generator},
    booktitle = {Proceedings of the IEEE/CVF Conference on Computer Vision and Pattern Recognition (CVPR)},
    month     = {June},
    year      = {2022},
    pages     = {3749-3761}
}

@inproceedings{NEURIPS2024_26cfdcd8,
	author = {Yang, Lihe and Kang, Bingyi and Huang, Zilong and Zhao, Zhen and Xu, Xiaogang and Feng, Jiashi and Zhao, Hengshuang},
	booktitle = {Advances in Neural Information Processing Systems},
	pages = {21875--21911},
	title = {Depth Anything V2},
	volume = {37},
	year = {2024}
}

@inproceedings{NEURIPS2020_d85b63ef,
 author = {Cubuk, Ekin Dogus and Zoph, Barret and Shlens, Jon and Le, Quoc},
 booktitle = {Advances in Neural Information Processing Systems},
 pages = {18613--18624},
 title = {RandAugment: Practical Automated Data Augmentation with a Reduced Search Space},
 volume = {33},
 year = {2020}
}

@inproceedings{Silberman:ECCV12,
  author    = {Nathan Silberman, Derek Hoiem, Pushmeet Kohli and Rob Fergus},
  title     = {Indoor Segmentation and Support Inference from RGBD Images},
  booktitle = {Proceedings of the European Conference on Computer Vision (ECCV)},
  pages={746--760},
  year      = {2012}
}

@inproceedings{zhou2020tracking,
  title={Tracking objects as points},
  author={Zhou, Xingyi and Koltun, Vladlen and Kr{\"a}henb{\"u}hl, Philipp},
  booktitle={Proceedings of the European Conference on Computer Vision (ECCV)},
  pages={474--490},
  year={2020},
}

@article{khanam2024yolov11,
  title={Yolov11: An overview of the key architectural enhancements},
  author={Khanam, Rahima and Hussain, Muhammad},
  journal={arXiv preprint arXiv:2410.17725},
  year={2024}
}

@InProceedings{Zhu_2019_CVPR,
author = {Zhu, Xizhou and Hu, Han and Lin, Stephen and Dai, Jifeng},
title = {Deformable ConvNets V2: More Deformable, Better Results},
booktitle = {Proceedings of the IEEE/CVF Conference on Computer Vision and Pattern Recognition (CVPR)},
month = {June},
year = {2019},
pages={9300-9308},
}

@InProceedings{Law_2018_ECCV,
author = {Law, Hei and Deng, Jia},
title = {CornerNet: Detecting Objects as Paired Keypoints},
booktitle = {Proceedings of the European Conference on Computer Vision (ECCV)},
month = {September},
pages={734-750},
year = {2018}
}

@inproceedings{bar2024egopet,
  title={Egopet: Egomotion and interaction data from an animal’s perspective},
  author={Bar, Amir and Bakhtiar, Arya and Tran, Danny and Loquercio, Antonio and Rajasegaran, Jathushan and LeCun, Yann and Globerson, Amir and Darrell, Trevor},
  booktitle={Proceedings of the European Conference on Computer Vision (ECCV)},
  pages={377--394},
  year={2024},
}

@inproceedings{habitat19iccv,
  title     =     {Habitat: {A} {P}latform for {E}mbodied {AI} {R}esearch},
  author    =     {{Manolis Savva*} and {Abhishek Kadian*} and {Oleksandr Maksymets*} and Yili Zhao and Erik Wijmans and Bhavana Jain and Julian Straub and Jia Liu and Vladlen Koltun and Jitendra Malik and Devi Parikh and Dhruv Batra},
  booktitle =     {Proceedings of the IEEE/CVF International Conference on Computer Vision (ICCV)},
  pages={9338-9346},
  year      =     {2019}
}

@InProceedings{Xia_2018_CVPR,
author = {Xia, Fei and Zamir, Amir R. and He, Zhiyang and Sax, Alexander and Malik, Jitendra and Savarese, Silvio},
title = {Gibson Env: Real-World Perception for Embodied Agents},
booktitle = {Proceedings of the IEEE/CVF Conference on Computer Vision and Pattern Recognition (CVPR)},
month = {June},
pages={9068-9079},
year = {2018}
}

@inproceedings{deng_imagenet_2009,
	title = {{ImageNet}: {A} large-scale hierarchical image database},
	shorttitle = {{ImageNet}},
	booktitle = {Proceedings of the IEEE/CVF Conference on Computer Vision and Pattern Recognition (CVPR)},
	author = {Deng, Jia and Dong, Wei and Socher, Richard and Li, Li-Jia and Li, Kai and Fei-Fei, Li},
	month = jun,
	year = {2009},
	pages = {248--255},
}

@inproceedings{lin_microsoft_2014,
	title = {Microsoft {COCO}: {Common} {Objects} in {Context}},
	booktitle = {Proceedings of the European Conference on Computer Vision (ECCV)},
	author = {Lin, Tsung-Yi and Maire, Michael and Belongie, Serge and Hays, James and Perona, Pietro and Ramanan, Deva and Dollár, Piotr and Zitnick, C. Lawrence},
	year = {2014},
	pages = {740--755},
}

@article{everingham_pascal_2010,
	title = {The {Pascal} {Visual} {Object} {Classes} ({VOC}) {Challenge}},
	volume = {88},
	number = {2},
	journal = {International Journal of Computer Vision},
	author = {Everingham, Mark and Van Gool, Luc and Williams, Christopher K. I. and Winn, John and Zisserman, Andrew},
	month = jun,
	year = {2010},
	pages = {303--338},
}

@InProceedings{Zhong_2025_ICCV,
    author    = {Zhong, Fangwei and Wu, Kui and Wang, Churan and Chen, Hao and Ci, Hai and Li, Zhoujun and Wang, Yizhou},
    title     = {UnrealZoo: Enriching Photo-realistic Virtual Worlds for Embodied AI},
    booktitle = {Proceedings of the IEEE/CVF International Conference on Computer Vision (ICCV)},
    month     = {October},
    year      = {2025},
    pages     = {5769-5779}
}

\end{document}